\documentclass[11pt]{article}

\usepackage[utf8]{inputenc}
\usepackage[T1]{fontenc}
\usepackage{amsmath,amssymb,amsthm}
\usepackage{mathtools}
\usepackage{booktabs}
\usepackage{graphicx}
\usepackage{xcolor}
\usepackage{algorithm}
\usepackage{algorithmic}
\usepackage{hyperref}
\usepackage{cleveref}
\usepackage[margin=1in]{geometry}
\usepackage{natbib}
\usepackage{thmtools}
\usepackage{listings}

\newtheorem{theorem}{Theorem}[section]
\newtheorem{lemma}[theorem]{Lemma}
\newtheorem{proposition}[theorem]{Proposition}
\newtheorem{corollary}[theorem]{Corollary}
\newtheorem{definition}[theorem]{Definition}
\newtheorem{remark}[theorem]{Remark}

\newcommand{\E}{\mathbb{E}}

\newcommand{\piroll}{\pi_{\mathrm{roll}}}
\newcommand{\pitheta}{\pi_{\theta}}

\newcommand{\DKL}{D_{\mathrm{KL}}}
\newcommand{\DTV}{D_{\mathrm{TV}}}
\newcommand{\Dklseq}{D_{\mathrm{KL}}^{\mathrm{seq}}}

\newcommand{\Dkltokmax}{D_{\mathrm{KL}}^{\mathrm{tok,max}}}
\newcommand{\Dtvtok}{D_{\mathrm{TV}}^{\mathrm{tok}}}
\newcommand{\Dtvtokmax}{D_{\mathrm{TV}}^{\mathrm{tok,max}}}
\newcommand{\Dtvseq}{D_{\mathrm{TV}}^{\mathrm{seq}}}

\newcommand{\Dbar}{\bar{D}}

\title{Trust Region Masking for Long-Horizon LLM Reinforcement Learning}

\usepackage[affil-it]{authblk}

\author{Yingru Li\thanks{Equal contribution}}
\author{Jiacai Liu\textsuperscript{*,1}}
\author{Jiawei Xu\textsuperscript{*,2}}
\author{Yuxuan Tong}
\author{Ziniu Li\textsuperscript{2}}
\author{\protect\\Qian Liu}
\author{Baoxiang Wang\textsuperscript{2}}

\affil{
\small ~~~~
\textsuperscript{1}Fudan University, \textsuperscript{2}The Chinese University of Hong Kong, Shenzhen
}
\date{}

\begin{document}

\maketitle

\begin{abstract}
Policy gradient methods for Large Language Models optimize a policy $\pi_\theta$ via a surrogate objective computed from samples of a rollout policy $\pi_{\text{roll}}$. However, modern LLM-RL pipelines suffer from unavoidable implementation divergences---backend discrepancies, Mixture-of-Experts routing discontinuities, and distributed training staleness---causing off-policy mismatch ($\pi_{\text{roll}} \neq \pi_\theta$) and approximation errors between the surrogate and the true objective.
We demonstrate that classical trust region bounds on this error scale as $O(T^2)$ with sequence length $T$, rendering them vacuous for long-horizon tasks. To address this, we derive a family of bounds---both KL-based and TV-based---including a \emph{Pinsker-Marginal} bound ($O(T^{3/2})$), a \emph{Mixed} bound ($O(T)$), and an \emph{Adaptive} bound that strictly generalizes the Pinsker-Marginal bound via per-position importance-ratio decomposition. Taking the minimum over all bounds yields the tightest known guarantee across all divergence regimes.
Crucially, all bounds depend on the maximum token-level divergence $\Dkltokmax$ (or $\Dtvtokmax$), a \emph{sequence-level} quantity that cannot be controlled by token-independent methods like PPO clipping. We propose Trust Region Masking (TRM), which masks entire sequences violating the trust region, enabling the first non-vacuous monotonic improvement guarantees for long-horizon LLM-RL.
\end{abstract}

\section{Introduction}
\label{sec:introduction}

Reinforcement Learning (RL) has become central to training Large Language Models (LLMs) for tasks requiring extended reasoning, multi-step problem solving, and agentic behavior~\citep{zeng2025simplerl, liu2024exploring, yang2024swe}. As response lengths expand from hundreds to thousands of tokens, policy gradient methods---particularly PPO~\citep{schulman2017proximal}---face increasingly strained theoretical foundations.

Trust region methods~\citep{kakade2002approximately, schulman2015trust} provide monotonic improvement guarantees by bounding the approximation error $|J(\pitheta) - J(\piroll) - L(\pitheta)|$ between the true objective $J$ and the surrogate $L$. This guarantee requires controlling the divergence between the rollout policy $\piroll$ and the training policy $\pitheta$---a divergence that is \emph{unavoidable} in modern LLM-RL due to three primary sources. \textbf{(i) Backend discrepancies:} high-throughput inference engines (vLLM~\citep{kwon2023efficient}, SGLang~\citep{zheng2024sglang}) use different attention kernels, precision formats, and operator fusion strategies than training frameworks (Megatron-LM~\citep{shoeybi2019megatron}, PyTorch FSDP~\citep{zhao2023pytorch}), causing logit differences that compound autoregressively. \textbf{(ii) MoE routing discontinuities:} in Mixture-of-Experts models~\citep{shazeer2017outrageously, liu2024deepseek}, minor numerical jitter can flip expert selection, causing high-magnitude jumps in token probabilities. \textbf{(iii) Distributed staleness:} asynchronous actor-learner architectures~\citep{espeholt2018impala, nair2015massively} introduce latency between data generation and gradient updates. These factors are documented in detail by~\citet{liu2025rlcollapse} and~\citet{yao2025offpolicy}; we provide further analysis in Appendix~\ref{app:mismatch}.

Classical trust region bounds~\citep{kakade2002approximately, achiam2017constrained} scale as $O(T^2)$ with sequence length. For a reasoning task with $T = 4096$ tokens and per-token divergence $\Dkltokmax = 10^{-4}$, the classical bound yields $|\mathrm{Error}| \le 1677$---vacuous given maximum reward of 1. While acceptable for short-horizon control tasks ($T \approx 100$), these bounds provide no guarantee that optimization actually improves performance at modern LLM scales.

Standard PPO addresses trust regions through token-level clipping~\citep{ziegler2019fine}, but autoregressive generation is inherently sequential: a small probability shift at an early token compounds through the entire trajectory, a phenomenon related to ``exposure bias''~\citep{bengio2015scheduled}. Distributed RL frameworks like IMPALA~\citep{espeholt2018impala} introduce correction mechanisms for staleness, but do not address the fundamental $O(T^2)$ scaling of the approximation error. Our work bridges this gap by deriving tight, non-vacuous bounds for autoregressive sequence generation and proposing sequence-level masking to enforce them.

We make the following contributions:
\begin{enumerate}
    \item \textbf{A family of tighter bounds (\Cref{sec:theory}):} We derive KL-based and TV-based variants of three bound families---Pinsker-Marginal ($O(T^{3/2})$), Mixed ($O(T)$), and Adaptive---and show that their minimum is the tightest known guarantee across all divergence regimes (\Cref{thm:unified}).

    \item \textbf{Trust Region Masking (\Cref{sec:trm}):} We show that all bounds depend on the maximum token-level divergence---a sequence-level quantity uncontrollable by token-level PPO clipping---and propose TRM, which masks entire sequences violating the trust region, enabling non-vacuous monotonic improvement guarantees. We demonstrate empirical training stability on mathematical reasoning benchmarks.
\end{enumerate}

\section{Background and Problem Setup}
\label{sec:background}

\subsection{Autoregressive Generation and Objective}

A policy $\pitheta$ generates a response $y = (y_1, \ldots, y_T)$ given prompt $x$, with trajectory probability $P^{\pitheta}(y \mid x) = \prod_{t=1}^T \pitheta(y_t \mid x, y_{<t})$. We define the context $c_t = (x, y_{<t})$ and the context visitation distribution:
\begin{equation}
    d_t^{\pi}(c_t) = P(x) \prod_{s=1}^{t-1} \pi(y_s \mid c_s).
\end{equation}
Given a scalar reward $R(x, y) \in [0, 1]$, the objective is $J(\pitheta) = \E_{x \sim P(x),\, y \sim \pitheta(\cdot|x)}[R(x, y)]$.

\subsection{The Surrogate Objective}

Samples are generated from a rollout policy $\piroll$ that generally differs from $\pitheta$. Following~\citet{kakade2002approximately}, we define the surrogate:
\begin{equation}
    L_{\piroll}(\pitheta) = \E_{\piroll}\!\left[ A \cdot \sum_{t=1}^T \rho_t \right],
    \label{eq:surrogate}
\end{equation}
where $A = R(x, y) - b$ is the trajectory advantage (with baseline $b$), $\rho_t = \pitheta(y_t \mid c_t) / \piroll(y_t \mid c_t)$ is the per-token importance ratio, and the gradient matches the true gradient at the reference: $\nabla L_{\piroll}(\pitheta)|_{\pitheta = \piroll} = \nabla J(\pitheta)|_{\pitheta = \piroll}$. The approximation error is:
\begin{equation}
    \mathrm{Error}(\pitheta) := J(\pitheta) - J(\piroll) - L(\pitheta).
    \label{eq:error-def}
\end{equation}
Bounding $|\mathrm{Error}|$ guarantees that maximizing $L$ leads to monotonic improvement in $J$.

\begin{remark}[Surrogate equivalence]
\label{rem:surrogate-equiv}
The error analysis uses the equivalent form $L'_{\piroll}(\pitheta) = \E_{\piroll}[R(x,y)\sum_{t}(\rho_t - 1)]$, which satisfies $L'_{\piroll}(\piroll) = 0$. The advantage form~\eqref{eq:surrogate} relates to it by $L_{\piroll}(\pitheta) = L'_{\piroll}(\pitheta) + T\,(J(\piroll) - b)$, a $\pitheta$-independent constant; the mean-reward baseline $b = J(\piroll)$ gives $L = L'$. Since $\nabla_\theta L = \nabla_\theta L'$ for any fixed $b$, the algorithmic gradient is unaffected by this choice. All value-level statements in this work (the conditions $\mathcal{M} > 0$ and $L > B^*$) are understood for the $L'$ form; equivalently, we set $b = J(\piroll)$ so that $L = L'$.
\end{remark}

\subsection{Divergence Measures}

We employ two families of divergence measures throughout.

\begin{definition}[Token-level divergences]
\label{def:token-div}
For context $c_t = (x, y_{<t})$:
\begin{align}
    \Dtvtok(c_t) &:= \DTV(\pitheta(\cdot|c_t) \| \piroll(\cdot|c_t)) = \tfrac{1}{2}\textstyle\sum_v |\pitheta(v|c_t) - \piroll(v|c_t)|, \\
    \DKL(c_t) &:= \DKL(\piroll(\cdot|c_t) \| \pitheta(\cdot|c_t)) = \textstyle\sum_v \piroll(v|c_t) \log\frac{\piroll(v|c_t)}{\pitheta(v|c_t)}.
\end{align}
\end{definition}

\begin{definition}[Sequence-level divergences]
\label{def:seq-div}
We define the maximum token-level divergences, the sequence-level divergences, and the expected per-position TV:
\begin{align}
    \epsilon &:= \Dtvtokmax := \max_{t, c_t} \Dtvtok(c_t), &
    \delta &:= \Dkltokmax := \max_{t, c_t} \DKL(c_t), \label{eq:eps-delta} \\
    \Dtvseq &:= \DTV(P^{\piroll}(\cdot|x) \| P^{\pitheta}(\cdot|x)), &
    \Dklseq &:= \textstyle\sum_{t=1}^T \E_{c_t \sim d_t^{\piroll}}[\DKL(c_t)], \label{eq:seq-divs} \\
    \Dbar_t &:= \E_{c_t \sim d_t^{\piroll}}\![\Dtvtok(c_t)]. & \label{eq:dbar}
\end{align}
\end{definition}

Throughout, $\delta, \epsilon, \Dklseq, \Dtvseq$, and $\Dbar_t$ are taken as worst-case (respectively, expected) over the prompt distribution $P(x)$ as well; the error bounds below hold pointwise in $x$ and therefore also after averaging over prompts. The key relationships between these quantities are (proofs in Appendix~\ref{app:lemma-proofs}):
\begin{itemize}
    \item \textbf{Pinsker's inequality:} $\Dtvtok(c_t) \le \min\!\big(1,\; \sqrt{\DKL(c_t)/2}\big)$, and therefore $\epsilon \le \min(1, \sqrt{\delta/2})$.
    \item \textbf{KL chain rule:} $\DKL(d_t^{\piroll} \| d_t^{\pitheta}) = \sum_{s=1}^{t-1}\E_{c_s \sim d_s^{\piroll}}[\DKL(c_s)] \le (t\!-\!1)\,\delta$.
    \item \textbf{Data processing inequality:} $\|d_t^{\pitheta} - d_t^{\piroll}\|_{\mathrm{TV}} \le \Dtvseq$.
\end{itemize}
Pinsker is \emph{tight} when the two distributions differ uniformly across the vocabulary, and \emph{loose} when divergence is concentrated on a few tokens (e.g., MoE routing flips). This motivates maintaining TV-based bounds alongside KL-based bounds: the KL route exploits sublinear context-shift scaling via Pinsker, while the TV route avoids Pinsker's looseness.

\section{Theoretical Analysis}
\label{sec:theory}

We derive a family of bounds on $|\mathrm{Error}|$ using both KL-based and TV-based routes. All proofs are in Appendix~\ref{app:proofs}.

\subsection{Error Decomposition and Building Blocks}
\label{subsec:building-blocks}

The error admits two exact decompositions, giving two routes to bounds; we develop the first here and the second in \Cref{subsec:adaptive}. The first is the Performance Difference Identity~\citep{kakade2002approximately}. Let $A_t^{\piroll}(c_t, y_t) := \E_{\piroll}[R \mid c_t, y_t] - \E_{\piroll}[R \mid c_t]$ be the per-step advantage and $g_t(c_t) := \E_{y_t \sim \pitheta}[A_t^{\piroll}(c_t, y_t)]$ the expected advantage shift. Then:
\begin{equation}
    \mathrm{Error} = \sum_{t=1}^T \Big( \E_{c_t \sim d_t^{\pitheta}}[g_t(c_t)] - \E_{c_t \sim d_t^{\piroll}}[g_t(c_t)] \Big).
    \label{eq:error-pdi}
\end{equation}
Each summand is bounded via $|\E_P[f] - \E_Q[f]| \le 2\|f\|_\infty \cdot \DTV(P, Q)$. Two building blocks control the factors:

\begin{lemma}[Advantage Bound]
\label{lem:advantage-bound}
For $R \in [0,1]$: $\|g_t\|_\infty \le 2\min\!\big(1,\;\epsilon,\;\sqrt{\delta/2}\big)$.
\end{lemma}

\begin{lemma}[Context Shift]
\label{lem:context-shift}
The context distribution shift is bounded by:
\begin{equation}
    \|d_t^{\pitheta} - d_t^{\piroll}\|_{\mathrm{TV}} \le \min\!\Big(1,\;\underbrace{(t\!-\!1)\,\epsilon}_{\text{coupling}},\;\underbrace{\sqrt{(t\!-\!1)\,\delta/2}}_{\text{Pinsker-on-marginal-KL}},\;\underbrace{\Dtvseq}_{\text{data processing}},\;\underbrace{\sqrt{\Dklseq/2}}_{\text{Pinsker-on-seq-KL}}\Big).
    \label{eq:context-shift-all}
\end{equation}
\end{lemma}

The classical bound takes the naive combination---worst-case advantage with the uncapped coupling shift---and is vacuous at scale.

\paragraph{The Classical Bound and Its Failure}

Using $\|g_t\|_\infty \le 2\epsilon$ with the coupling bound $(t\!-\!1)\epsilon$ and no caps:
\begin{equation}
    |\mathrm{Error}| \le 4\epsilon^2 \sum_{t=1}^T (t\!-\!1) = 2T(T\!-\!1)\epsilon^2.
    \label{eq:classical}
\end{equation}
Since $\epsilon \le \sqrt{\delta/2}$, this implies $|\mathrm{Error}| \le T(T\!-\!1)\delta$. At $T = 4096$, $\delta = 10^{-4}$: $|\mathrm{Error}| \le 4096 \times 4095 \times 10^{-4} \approx 1677$---\emph{vacuous}.

\subsection{Worst-Case Bounds}
\label{subsec:worst-case}

These bounds follow the first route: each pairs the advantage factor of \Cref{lem:advantage-bound} with one context-shift term of \Cref{lem:context-shift}, and depends only on the scalar divergences $\delta,\epsilon,\Dklseq,\Dtvseq$.

\paragraph{Pinsker-Marginal Bounds}

Applying Pinsker's inequality to the \emph{marginal} KL gives a sublinear context shift $\sqrt{(t\!-\!1)\delta/2}$, while the advantage factor can use either the KL-derived bound $\sqrt{\delta/2}$ or the direct TV value $\epsilon$:

\begin{theorem}[Pinsker-Marginal Bounds]
\label{thm:pinsker-marginal}
\begin{align}
    B_{\mathrm{PM}}^{\mathrm{KL}} &= 4\min\!\Big(1,\sqrt{\tfrac{\delta}{2}}\Big) \sum_{t=1}^{T}\min\!\Big(1, \sqrt{\tfrac{(t-1)\delta}{2}}\Big), \label{eq:B-PM-KL} \\
    B_{\mathrm{PM}}^{\mathrm{TV}} &= 4\,\min(1,\epsilon) \sum_{t=1}^{T}\min\!\Big(1, \sqrt{\tfrac{(t-1)\delta}{2}}\Big). \label{eq:B-PM-TV}
\end{align}
In the small-divergence regime ($\delta \le 2/T$, so all caps are inactive):
\begin{equation}
    B_{\mathrm{PM}}^{\mathrm{KL}} = \tfrac{4}{3}T^{3/2}\delta, \qquad
    B_{\mathrm{PM}}^{\mathrm{TV}} = \tfrac{8}{3}T^{3/2}\,\epsilon\sqrt{\delta/2}.
    \label{eq:PM-simplified}
\end{equation}
\end{theorem}

The TV variant replaces $\sqrt{\delta/2}$ with $\epsilon$ in the advantage factor, gaining whenever Pinsker is loose ($\epsilon < \sqrt{\delta/2}$). When Pinsker is tight ($\epsilon = \sqrt{\delta/2}$), both variants coincide.

\paragraph{Mixed Bounds}

Using a \emph{uniform} context-shift bound from the sequence-level divergence (which does not grow with $t$):

\begin{theorem}[Mixed Bounds]
\label{thm:mixed}
\begin{align}
    B_{\mathrm{Mix}}^{\mathrm{KL}} &= 4T \cdot \min\!\Big(1, \sqrt{\tfrac{\delta}{2}}\Big) \cdot \min\!\Big(1, \sqrt{\tfrac{\Dklseq}{2}}\Big), \label{eq:B-Mix-KL} \\
    B_{\mathrm{Mix}}^{\mathrm{TV}} &= 4T \cdot \min(1,\,\epsilon) \cdot \min(1,\, \Dtvseq). \label{eq:B-Mix-TV}
\end{align}
When both caps are inactive: $B_{\mathrm{Mix}}^{\mathrm{KL}} = 2T\sqrt{\delta \cdot \Dklseq}$ and $B_{\mathrm{Mix}}^{\mathrm{TV}} = 4T\,\epsilon\,\Dtvseq$.
\end{theorem}

The Mixed bounds scale as $O(T)$ and are tighter when divergence is sparse ($\Dklseq \ll T\delta$).

\paragraph{Coupling Bounds}

The pure TV-coupling route avoids all KL/Pinsker conversions:

\begin{theorem}[Coupling Bound]
\label{thm:coupling}
\begin{equation}
    B_{\mathrm{Coup}} = 4\,\min(1,\epsilon) \sum_{t=1}^{T}\min\!\big(1,\;(t\!-\!1)\,\epsilon\big).
    \label{eq:B-Coup}
\end{equation}
When $T\epsilon \le 1$: $B_{\mathrm{Coup}} = 2T(T\!-\!1)\epsilon^2$ (classical). When $T\epsilon > 1$: $B_{\mathrm{Coup}} \le 4T\epsilon$ ($O(T)$ via the cap).
\end{theorem}

\subsection{The Data-Dependent (Adaptive) Bound}
\label{subsec:adaptive}

Following the second route, the Adaptive bound starts from the importance-ratio decomposition $J(\pitheta) - J(\piroll) = L'(\pitheta) - \Delta$, with the remainder $\Delta$ given in \Cref{lem:exact-identity}, and a tower-property factorization (\Cref{app:proof-adaptive}); unlike the worst-case bounds it retains the per-position divergence $\Dbar_t$ rather than its maximum.

\begin{theorem}[Adaptive Bounds]
\label{thm:adaptive}
\begin{align}
    B_{\mathrm{Adap}}^{\mathrm{KL}} &= 4 \sum_{t=1}^{T} \Dbar_t \cdot \min\!\Big(1,\; \sqrt{\tfrac{(T-t)\,\delta}{2}}\Big), \label{eq:B-Adap-KL} \\
    B_{\mathrm{Adap}}^{\mathrm{TV}} &= 4 \sum_{t=1}^{T} \Dbar_t \cdot \min\!\big(1,\;(T\!-\!t)\,\epsilon\big), \label{eq:B-Adap-TV} \\
    B_{\mathrm{Adap}}^{*} &= 4 \sum_{t=1}^{T} \Dbar_t \cdot \min\!\Big(1,\;(T\!-\!t)\,\epsilon,\;\sqrt{\tfrac{(T-t)\,\delta}{2}}\Big). \label{eq:B-Adap-star}
\end{align}
The hybrid bound $B_{\mathrm{Adap}}^{*} \le \min(B_{\mathrm{Adap}}^{\mathrm{KL}}, B_{\mathrm{Adap}}^{\mathrm{TV}})$ by construction.
\end{theorem}

The Adaptive bounds are tighter than the Pinsker-Marginal and Coupling bounds in two independent ways:
\begin{enumerate}
    \item \textbf{Data-dependent advantage:} $\Dbar_t$ (expected per-position TV under $\piroll$) replaces $\epsilon$ or $\sqrt{\delta/2}$ (worst-case). When divergence is non-uniform---e.g., concentrated at a few tokens due to MoE routing---$\Dbar_t \ll \epsilon$ at most positions.
    \item \textbf{Adaptive future bounding:} The future-trajectory TV is bounded by the tighter of two routes at each position. The Pinsker route gives $\sqrt{(T\!-\!t)\delta/2}$ (sublinear in remaining horizon); the coupling route gives $(T\!-\!t)\epsilon$ (linear but avoids Pinsker looseness). The crossover occurs at $T - t = \delta/(2\epsilon^2)$.
\end{enumerate}

\begin{remark}[Recovery of prior bounds]
\label{rem:relationship}
The worst-case substitutions $\Dbar_t \mapsto \min(1,\sqrt{\delta/2})$ and $\Dbar_t \mapsto \epsilon$ recover $B_{\mathrm{PM}}^{\mathrm{KL}}$ and $B_{\mathrm{Coup}}$ exactly, both via the index reversal $\sum_t f(T\!-\!t) = \sum_t f(t\!-\!1)$ (\Cref{app:proof-adaptive}, Step 5). Since $\Dbar_t \le \min(1,\epsilon,\sqrt{\delta/2})$ pointwise, the Adaptive family is at least as tight as $B_{\mathrm{PM}}^{\mathrm{KL}}$, $B_{\mathrm{PM}}^{\mathrm{TV}}$, and $B_{\mathrm{Coup}}$, and strictly tighter when divergence is non-uniform across positions.
\end{remark}

\subsection{The Unified Bound}
\label{sec:unified}

Since all bounds hold independently, their minimum is valid:

\begin{theorem}[Unified Bound]
\label{thm:unified}
The approximation error satisfies $|\mathrm{Error}| \le B^*$, where:
\begin{equation}
    \boxed{B^* = \min\!\Big\{ B_{\mathrm{PM}}^{\mathrm{KL}},\; B_{\mathrm{PM}}^{\mathrm{TV}},\; B_{\mathrm{Mix}}^{\mathrm{KL}},\; B_{\mathrm{Mix}}^{\mathrm{TV}},\; B_{\mathrm{Coup}},\; B_{\mathrm{Adap}}^{*}\Big\}.}
    \label{eq:unified-bound}
\end{equation}
Defining $\mathcal{M}(\pitheta) := L(\pitheta) - B^*$, the condition $\mathcal{M}(\pitheta) > 0$ guarantees monotonic improvement: $J(\pitheta) > J(\piroll)$.
\end{theorem}

\begin{remark}[The adaptive bound subsumes the marginal and coupling routes]
\label{rem:adap-dominates}
By \Cref{rem:relationship}, when the $\Dbar_t$ are available $B_{\mathrm{Adap}}^{*} \le \min\{B_{\mathrm{PM}}^{\mathrm{KL}}, B_{\mathrm{PM}}^{\mathrm{TV}}, B_{\mathrm{Coup}}\}$: the Adaptive family already subsumes the marginal and coupling routes. The operative minimum is then $\min\{B_{\mathrm{Mix}}^{\mathrm{KL}}, B_{\mathrm{Mix}}^{\mathrm{TV}}, B_{\mathrm{Adap}}^{*}\}$---the Mixed route is the only complement, as it alone uses the aggregate divergences $\Dklseq, \Dtvseq$ rather than the maxima $\delta, \epsilon$. The other three are retained in~\eqref{eq:unified-bound} as the computable fallback when $\Dbar_t$ cannot be estimated.
\end{remark}

\begin{table*}[t]
\centering
\caption{%
  Error bounds for $T = 4096$.
  \textbf{Small divergence:} $\delta = 10^{-4}$, $\epsilon = 5 \times 10^{-3}$ (actual TV $< \sqrt{\delta/2} \approx 7.07 \times 10^{-3}$, so Pinsker is loose), $\Dklseq = 0.01$, $\Dtvseq = 0.05$.
  \textbf{KL-only:} same $\delta$ and $\Dklseq$, with $\epsilon$ and $\Dtvseq$ derived via Pinsker ($\epsilon = 7.07 \times 10^{-3}$, $\Dtvseq = 0.0707$), showing that TV bounds reduce to KL bounds.
  The unified bound $B^*$ (bottom row) takes the minimum over all rows.%
}
\label{tab:bounds}
\renewcommand{\arraystretch}{1.3}
\begin{tabular}{l l c r r}
\toprule
\textbf{Bound} & \textbf{Route} & \textbf{Scaling} &
  \textbf{KL-only} & \textbf{KL+TV} \\
\midrule
Classical & TV (uncapped) & $O(T^{2})$ & $1{,}677$ & $839$ \\[3pt]
Coupling & TV coupling (capped) & $O(T)^{\dagger}$ & $113.8$ & $79.9$ \\[3pt]
Pinsker-Marginal & KL & $O(T^{3/2})$ & $35.0$ & $35.0$ \\
Pinsker-Marginal & TV+KL & $O(T^{3/2})$ & $35.0$ & $24.7$ \\[3pt]
Mixed & KL & $O(T)$ & $8.2$ & $8.2$ \\
Mixed & TV & $O(T)$ & $8.2$ & $4.1$ \\[3pt]
Adaptive (hybrid) & KL+TV & data-dep. & $\le 35.0^{*}$ & $\le 24.7^{*}$ \\
\midrule
\textbf{Unified} $B^*$ & \textbf{min all} & --- & $\mathbf{\le 8.2}$ & $\mathbf{\le 4.1}$ \\
\bottomrule
\end{tabular}

\smallskip
{\footnotesize
$^{\dagger}$\,$O(T^2)$ without caps; $O(T)$ when $T\epsilon > 1$ and the $\min(1, \cdot)$ cap activates.\quad
$^{*}$\,Worst-case Adaptive hybrid (uniform $\Dbar_t$): reduces to PM-KL when $\epsilon = \sqrt{\delta/2}$, and to PM-TV when $\epsilon < \sqrt{\delta/2}$. Strictly tighter with non-uniform $\Dbar_t$.}
\end{table*}

\textbf{All bounds depend on token-level maxima.} Whether parameterized by $\delta = \Dkltokmax$ or $\epsilon = \Dtvtokmax$, every bound depends on the worst-case per-token divergence. Controlling only the average is provably insufficient:

\begin{proposition}
\label{prop:no-pure-seq}
There exists no function $f\!: \mathbb{R}^+ \to \mathbb{R}^+$ such that $\Dkltokmax \le f(\Dklseq)$ for all policy pairs.
\end{proposition}
\begin{proof}
Let $c^*$ occur with probability $p$ under $\piroll$, with $\DKL(c^*) = 1$ and $\DKL(c) = 0$ for $c \neq c^*$. Then $\Dkltokmax = 1$ while $\Dklseq = p \to 0$ as $p \to 0$.
\end{proof}

\textbf{KL and TV routes are complementary.} The KL route gives sublinear context-shift scaling but a lossy advantage factor; the TV route gives a tight advantage factor but only linear context-shift scaling. Neither dominates, and the hybrid $B_{\mathrm{Adap}}^*$ selects the tighter route per position (\Cref{subsec:adaptive}). In every case the bound is governed by the token-level \emph{maximum}---which, as the next section shows, is precisely what token-level methods cannot control.

\section{Trust Region Masking}
\label{sec:trm}

\subsection{Why Token-Level Methods Fail}
\label{subsec:token-failure}

\Cref{sec:theory} established that every bound is governed by the token-level maximum $\Dkltokmax$, an irreducible sequence-level quantity (\Cref{prop:no-pure-seq}). We now show that the two standard token-level interventions cannot control it, for two structural reasons.

\textbf{PPO clipping leaks gradients.} PPO~\citep{schulman2017proximal} constrains updates via clipping: $L^{\mathrm{CLIP}} = \E[\sum_t \min(\rho_t A_t, \mathrm{clip}(\rho_t, 1\!-\!\epsilon_{\mathrm{clip}}, 1\!+\!\epsilon_{\mathrm{clip}}) A_t)]$, where $\epsilon_{\mathrm{clip}}$ is the clip width (distinct from the divergence $\epsilon = \Dtvtokmax$). The clipping is asymmetric: when $\rho_t \gg 1 + \epsilon_{\mathrm{clip}}$ and $A_t < 0$, the objective uses the unclipped $\rho_t A_t$. In standard RL, this correctly penalizes over-represented bad actions. Under implementation divergence, the large $\rho_t$ reflects numerical noise (e.g., MoE routing flip where $\piroll(v) = 0.9$ but $\pitheta(v) = 0.001$, giving $\rho \approx 900$), producing a massive erroneous gradient.

\textbf{Token masking preserves vacuous bounds.} Zeroing the gradient of outlier tokens (where $|\log \rho_t| > \tau$) prevents immediate gradient explosion but does not change the underlying divergence $\Dkltokmax$ between $\piroll$ and $\pitheta$. The trust region remains violated:

\begin{remark}[Token masking does not restore the guarantee]
\label{rem:token-masking-fails}
Token masking acts on the \emph{loss}: it zeroes selected per-token gradient terms but leaves the policy pair $(\piroll, \pitheta)$ unchanged. Since $\Dkltokmax$ is a property of that pair---not of which terms enter the loss---it is unaffected, and the precondition $\Dkltokmax \le \delta$ underlying \Cref{sec:theory} remains violated. Token masking therefore alters the optimization direction without restoring the monotonic-improvement guarantee.
\end{remark}

The root cause is that the approximation error is cumulative over the sequence. If \emph{any} token violates the trust region, the entire trajectory is compromised as an estimator for $J(\pitheta)$, and the entire sequence must be excluded.

\subsection{The Masked Surrogate Objective}

We define a binary sequence mask $M(x, y) = \mathbb{I}[(x, y) \in \text{Trust Region}]$ and the masked surrogate:
\begin{equation}
    L_{\mathrm{masked}}(\pitheta) = \E_{\piroll}\!\left[ M(x, y) \cdot A(x, y) \cdot \sum_{t=1}^T \rho_t \right].
    \label{eq:masked-surrogate}
\end{equation}
The gradient is normalized by the \emph{total} batch size $N$ (not the accepted count), so rejected sequences contribute zero gradient. This is a rejection sampling mechanism: we choose not to learn from trajectories where off-policy divergence renders the gradient unreliable.

\subsection{Masking Criterion and Implementation}

\textbf{Exact KL computation.} Because $\piroll$ logits are stored during rollout and $\pitheta$ logits are computed during the training forward pass, $\DKL(c_t)$ is computed exactly---over the full vocabulary---at no extra inference cost:
\begin{equation}
    \DKL(c_t) = \sum_{v \in \mathcal{V}} \piroll(v|c_t) \log \frac{\piroll(v|c_t)}{\pitheta(v|c_t)}.
\end{equation}

\textbf{Max-based criterion:} $M(x,y) = \mathbb{I}[\max_t \DKL(c_t) \le \delta]$, directly bounding $\Dkltokmax$. The threshold $\delta$ is \emph{length-invariant}: it does not depend on $T$. Because both $\piroll$ and $\pitheta$ logits are available, $\Dtvtok(c_t)$ can be computed alongside $\DKL(c_t)$; masking instead (or additionally) on $\max_t \Dtvtok(c_t) \le \epsilon$ certifies a small $\Dtvtokmax$ directly, unlocking the tighter TV-route bounds---which otherwise collapse to their KL counterparts, since a KL-only criterion certifies only $\epsilon \le \sqrt{\delta/2}$.

\textbf{Combined criterion:} In practice, combining max and average constraints ($\frac{1}{T}\sum_t \DKL(c_t) \le \delta_{\mathrm{avg}}$) allows looser individual thresholds while maintaining robustness.

\textbf{Sample-based approximation.} When full logits are unavailable, we recommend $k_2(\rho) = \frac{1}{2}(\log \rho)^2$ for max-filtering (symmetric, detects both $\rho \to 0$ and $\rho \to \infty$) and $k_3(\rho) = \rho - 1 - \log \rho$ for averaging (unbiased: $\E_{\piroll}[k_3] = \DKL(\piroll \| \pitheta)$). See Appendix~\ref{app:k3}.

\begin{algorithm}[t]
\caption{Trust Region Masking (TRM)}
\label{alg:trm}
\begin{algorithmic}[1]
    \REQUIRE Threshold $\delta$; Batch $\mathcal{D} = \{(x^{(i)}, y^{(i)})\}_{i=1}^N$; Stored $\piroll$ logits
    \STATE \textbf{Forward Pass:} Compute $\pitheta$ logits on all data
    \FOR{each sequence $i$}
        \STATE Compute $\DKL(c_t^{(i)})$ for all $t$ using stored $\piroll$ and current $\pitheta$ logits
        \STATE $M_i \leftarrow \mathbb{I}[\max_{t} \DKL(c_t^{(i)}) \le \delta]$
    \ENDFOR
    \STATE \textbf{Backward Pass:} $\nabla L_{\mathrm{masked}} \leftarrow \frac{1}{N}\sum_{i=1}^{N} M_i \cdot A^{(i)} \cdot \sum_{t} \rho_t^{(i)} \nabla\log\pitheta(y_t^{(i)}|c_t^{(i)})$
    \STATE \textbf{Update:} $\theta \leftarrow \theta + \alpha \cdot \nabla L_{\mathrm{masked}}$
\end{algorithmic}
\end{algorithm}

\subsection{Guarantee Under the Global Precondition}

\begin{theorem}[TRM Guarantee]
\label{thm:trm-guarantee}
Algorithm~\ref{alg:trm} with threshold $\delta$ satisfies:
\begin{enumerate}
    \item \textbf{Bounded divergence:} For all accepted sequences ($M_i = 1$), $\max_t \DKL(c_t^{(i)}) \le \delta$.
    \item \textbf{Length-invariant threshold:} $\delta$ does not depend on sequence length $T$.
    \item \textbf{Non-vacuous error bound:} If additionally $\Dkltokmax \le \delta$ holds globally (for all reachable contexts), then:
    \begin{equation}
        |J(\pitheta) - J(\piroll) - L(\pitheta)| \le B^*(\delta, \epsilon),
    \end{equation}
    where $B^*$ is the unified bound (\Cref{thm:unified}) with $\Dkltokmax$ replaced by $\delta$. The condition $L(\pitheta) > B^*$ guarantees $J(\pitheta) > J(\piroll)$.
\end{enumerate}
\end{theorem}

\begin{proof}
Part~(1) holds by construction of the masking criterion. Part~(2) follows because the criterion $\max_t \DKL(c_t) \le \delta$ depends only on the per-token maximum, not on $T$. Part~(3): when $\Dkltokmax \le \delta$ globally, all bounds in \Cref{sec:theory} apply with $\Dkltokmax$ replaced by $\delta$, giving $|\mathrm{Error}| \le B^*$. Since $J(\pitheta) - J(\piroll) = L(\pitheta) + \mathrm{Error} \ge L(\pitheta) - B^*$, the condition $L(\pitheta) > B^*$ ensures $J(\pitheta) > J(\piroll)$.
\end{proof}

\begin{remark}[Role of masking in the guarantee]
\label{rem:masking-role}
The error bound involves the \emph{full} surrogate $L$, whereas TRM optimizes the \emph{masked} surrogate $L_{\mathrm{masked}}$. These are related by a sharp dichotomy, not an approximation:
\begin{itemize}
    \item \textbf{Precondition holds} ($\Dkltokmax \le \delta$ for all reachable contexts). Then every $y \in \mathrm{supp}(\piroll)$ satisfies $\max_t \DKL(c_t) \le \delta$, so $M \equiv 1$ $\piroll$-almost everywhere and $L_{\mathrm{masked}} = L$ \emph{exactly}. The same identity makes the stop-gradient mask exact ($\nabla M = 0$) and validates $B^*$. Masking is then redundant on-support: it removes nothing, but certifies that nothing needed removing.
    \item \textbf{Precondition fails.} Some sampled sequences are rejected and $L_{\mathrm{masked}} \neq L$. A high acceptance rate does \emph{not} bound $|L - L_{\mathrm{masked}}|$: rejected sequences carry disproportionate importance-sampling mass (a large $\rho_t$ is precisely what triggers rejection), so $|L - L_{\mathrm{masked}}|$ can be $\Theta(\|A\|_\infty T)$ even as the rejection rate $\to 0$. In this regime $B^*$ is also invalid, so the entire right-hand side of \Cref{thm:trm-guarantee}(3) is moot.
\end{itemize}
All three gaps---population ($L$ vs.\ $L_{\mathrm{masked}}$), gradient (stop-gradient vs.\ true $\nabla L_{\mathrm{masked}}$), and validity of $B^*$---open and close together at $M \equiv 1$. Masking therefore does \emph{not} close a population-level gap. Its two honest roles are: \textbf{(1) Falsification:} the empirical acceptance rate tests the precondition---a rate near $1$ is evidence that $\Dkltokmax \le \delta$ holds globally, while a low rate \emph{refutes} it and signals that $B^*$ does not apply; \textbf{(2) Estimator robustness:} discarding high-$\rho_t$ trajectories removes the dominant variance and the destructive updates of \Cref{subsec:token-failure} from the \emph{finite-sample} gradient, independently of the population identity. We recommend monitoring acceptance above $70\%$ as a precondition diagnostic, not as a proxy for $L_{\mathrm{masked}} \approx L$.
\end{remark}

\paragraph{Numerical illustration.} At $T = 4096$, $\delta = 10^{-4}$, $\epsilon = 5 \times 10^{-3}$, $\Dklseq = 0.01$, $\Dtvseq = 0.05$: the unified bound gives $B^* \le 4.1$ (TV-Mixed), versus the classical $1677$---a $\mathbf{409\times}$ improvement.

\paragraph{Length bias.} The threshold $\delta$ is length-invariant, but the \emph{probability of rejection} grows with $T$: a longer sequence has more tokens that can individually violate the criterion. This systematically penalizes long reasoning chains. Appendix~\ref{app:length-neutral} develops length-neutral variants (LN-TRM, SER) that normalize the per-token error and remove this bias while preserving the guarantee up to a controlled factor.

\subsection{A Precondition-Free Guarantee}
\label{subsec:precond-free}

\Cref{rem:masking-role} shows that $B^*$ cannot be transferred from the full surrogate $L$ to the masked surrogate $L_{\mathrm{masked}}$. A guarantee for the masked objective is nonetheless available \emph{without} the global precondition: decomposing the performance difference over the accepted and rejected regions gives $J(\pitheta) - J(\piroll) \ge L'_{\mathrm{masked}}(\pitheta) - B_{\mathcal{A}} - (r + \Dtvseq)$, where $L'_{\mathrm{masked}}$ is the masked surrogate ($R$-form), $r = \Pr_{\piroll}[\text{reject}]$ is the empirical masking rate, and $B_{\mathcal{A}}$ bounds the error on the accepted region. The rejection rate thus enters \emph{linearly and observably}, formalizing the recommendation to monitor it. This is a deliberate trade-off: the global precondition of \Cref{thm:trm-guarantee} is what buys the sublinear-in-$T$ scaling of \Cref{sec:theory}, which the precondition-free bound forgoes---taking $B_{\mathcal{A}} = B^*$ recovers \Cref{thm:trm-guarantee} exactly under the precondition, whereas a pointwise sample criterion yields a precondition-free $B_{\mathcal{A}}$ at an exponential-in-$\tau$ cost. \Cref{app:precond-free} gives the full statement, proof, and both regimes.

\section{Experiments}
\label{sec:experiments}

We validate TRM on mathematical reasoning using Qwen3-8B-Base under Zero-RL setup~\citep{guo2025deepseek}, trained on deduplicated DAPO-MATH-17k and evaluated on AIME25. We use GRPO~\citep{shao2024deepseekmath} with group size 16, batch size 32, and learning rate $1 \times 10^{-6}$. Evaluation uses Top-P$=0.95$, Temperature$=1.0$, reporting avg@32.

To simulate realistic mismatch, we use vLLM for inference and PyTorch FSDP for training. We measure the \emph{Log Absolute Perplexity (PPL) Gap}:
\begin{equation}
    \Delta_{\mathrm{PPL}} = \frac{1}{N} \sum_{i=1}^N \left| \frac{1}{T_i} \sum_{t=1}^{T_i} \log \pitheta(y_t^{(i)} | c_t^{(i)}) - \frac{1}{T_i}\sum_{t=1}^{T_i}\log \piroll(y_t^{(i)} | c_t^{(i)}) \right|.
\end{equation}

\begin{figure}[htbp]
    \centering
    \includegraphics[width=0.65\linewidth]{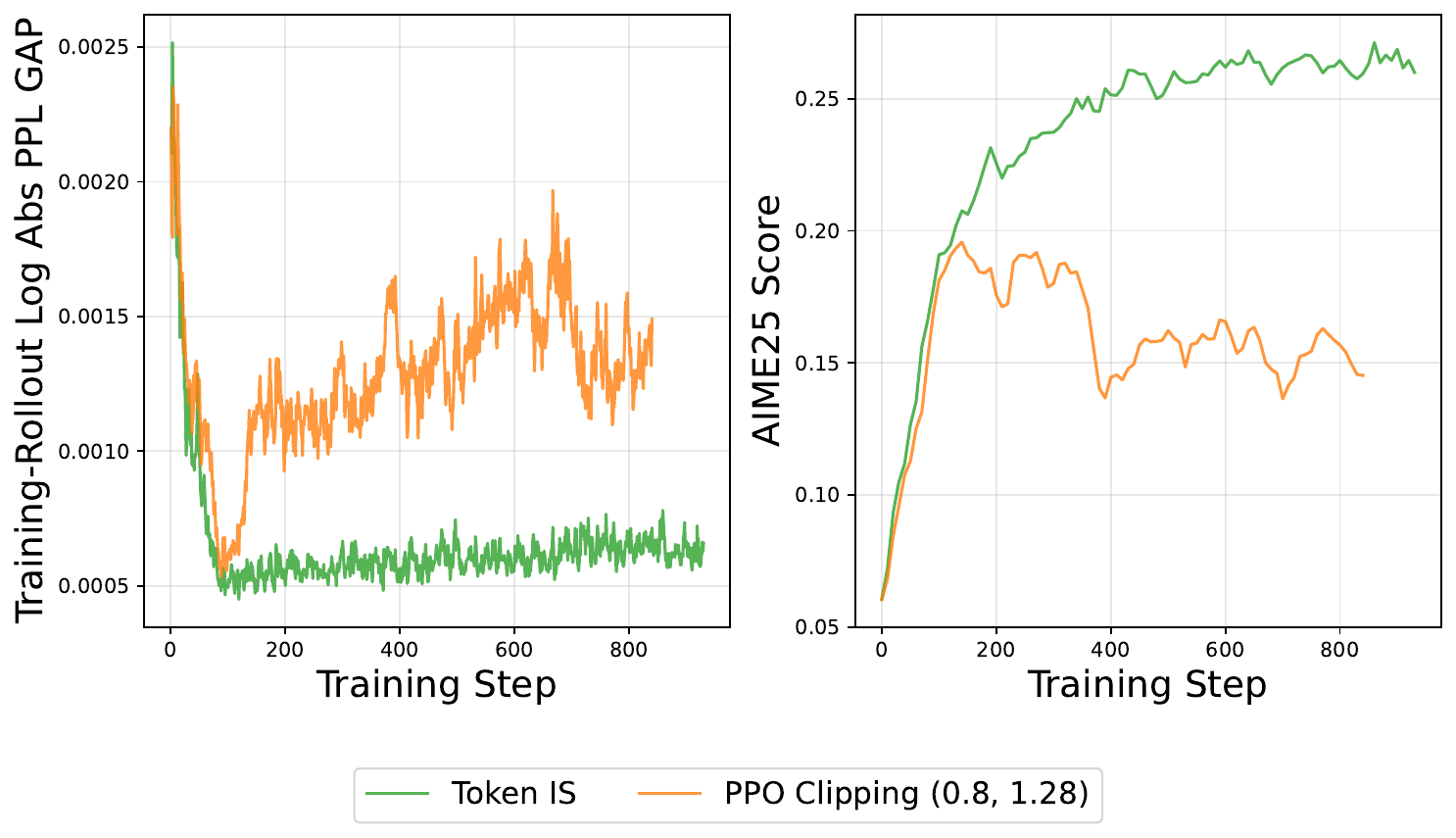}
    \caption{Token-level IS vs.\ PPO Clipping. Clipping exacerbates instability (larger PPL Gap, degraded score), confirming that token-level interventions cannot control $\Dkltokmax$.}
    \label{fig:vs_tokenis}
\end{figure}

\begin{figure}[htbp]
    \centering
    \includegraphics[width=0.65\linewidth]{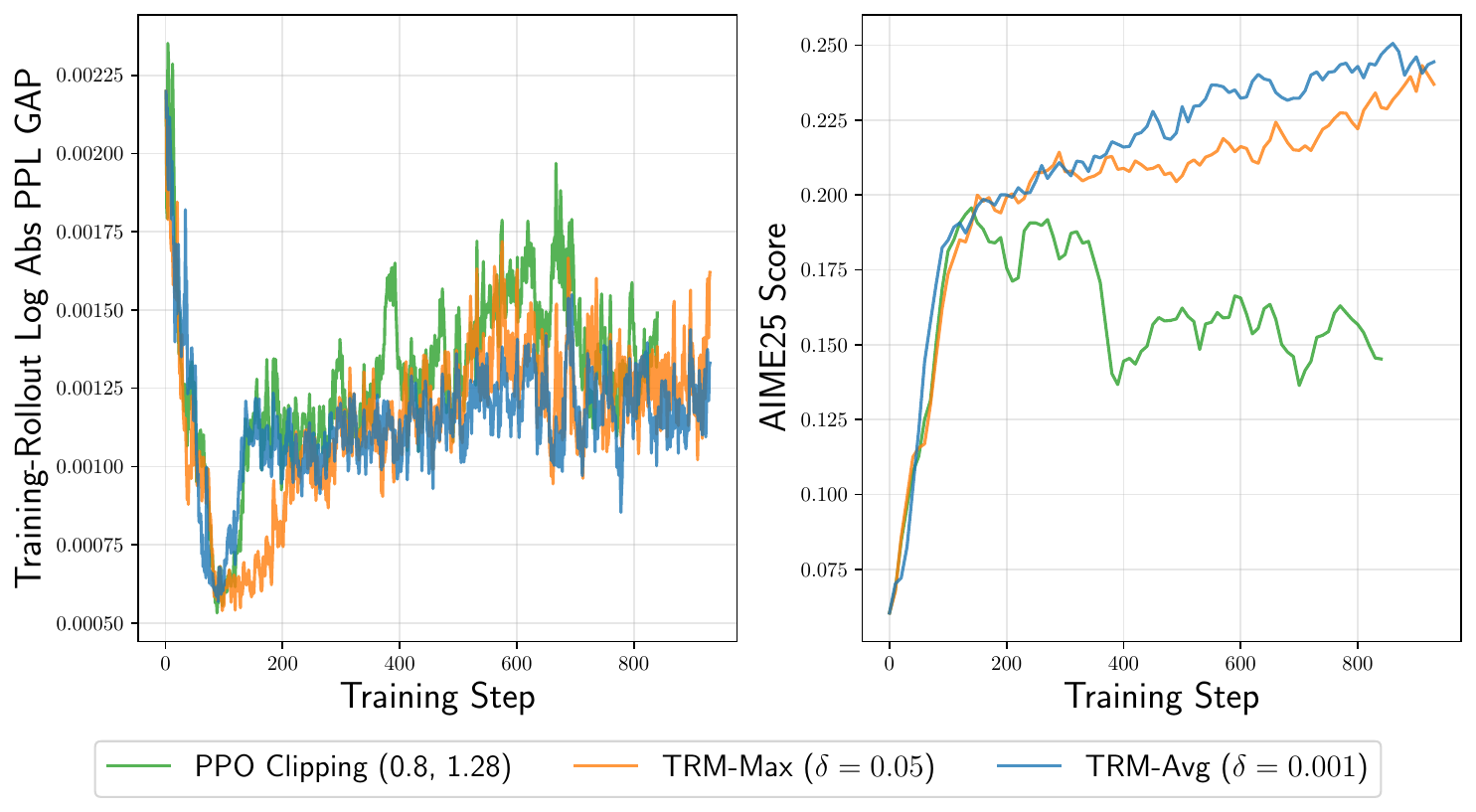}
    \caption{TRM vs.\ PPO Clipping. Both TRM-Max ($\delta\!=\!0.05$) and TRM-Avg ($\delta\!=\!0.001$) stabilize training, keeping the PPL Gap bounded.}
    \label{fig:vs_ppo_clip}
\end{figure}

\Cref{fig:vs_tokenis} confirms that token-level PPO clipping exacerbates instability. \Cref{fig:vs_ppo_clip} shows that both TRM variants maintain stability and consistent improvement on AIME25, keeping the PPL Gap bounded by rejecting mismatched sequences.

\begin{figure}[htbp]
    \centering
    \includegraphics[width=0.65\linewidth]{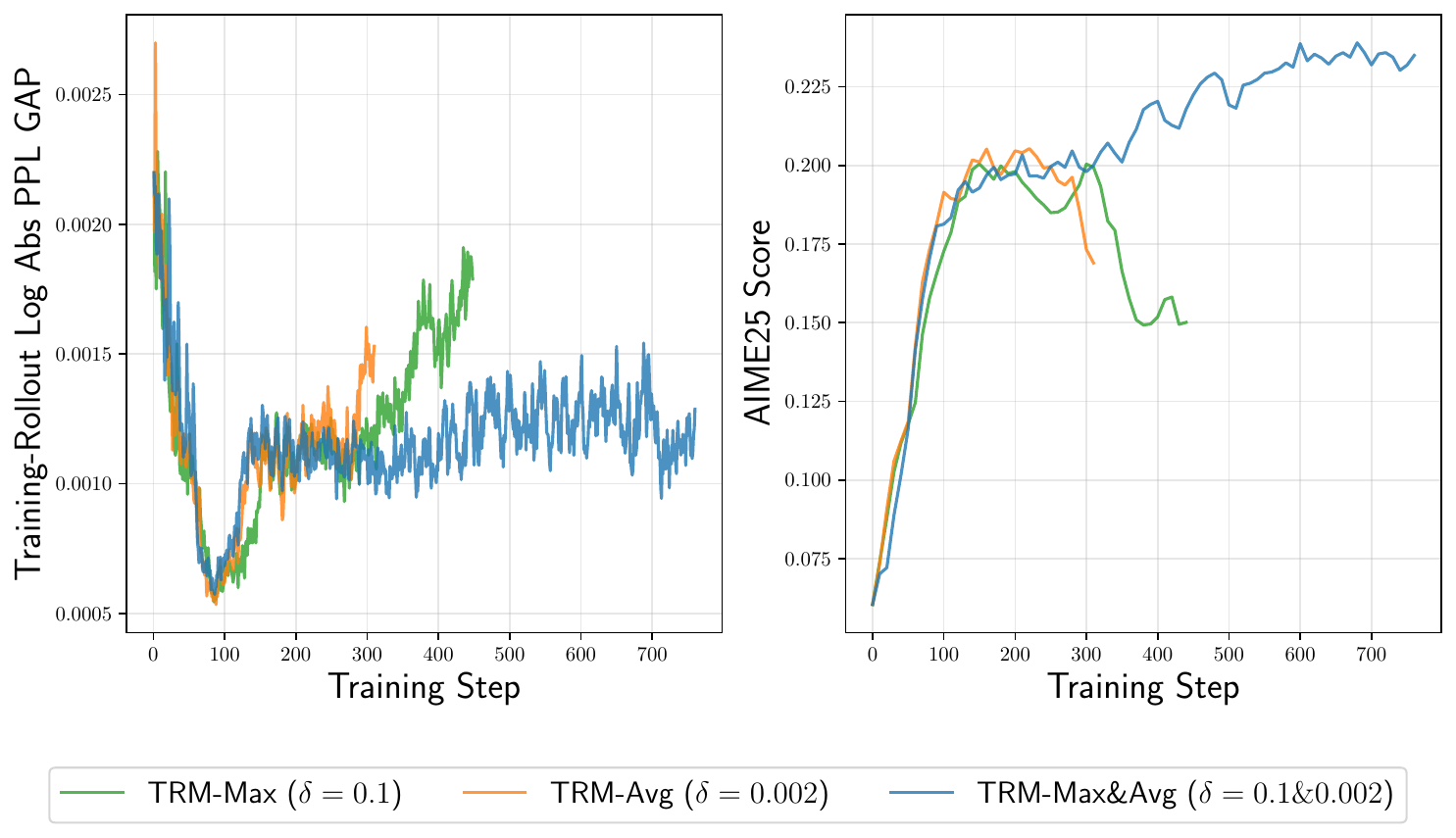}
    \caption{Combined criteria. Individually loose TRM-Max ($\delta\!=\!0.1$) and TRM-Avg ($\delta\!=\!0.002$) fail, but their combination succeeds---the max criterion catches outliers while the average criterion limits accumulated drift.}
    \label{fig:combined_criterion}
\end{figure}

\section{Conclusion}
\label{sec:conclusion}

Off-policy mismatch is unavoidable in modern LLM-RL. We show that classical $O(T^2)$ trust region bounds are vacuous for long-horizon tasks and derive a family of tighter bounds---both KL-based and TV-based---whose minimum yields the tightest known guarantee. The KL route provides sublinear context-shift scaling; the TV route avoids Pinsker's looseness. Their combination strictly dominates either alone. All bounds depend on the maximum token-level divergence, which cannot be controlled by token-level methods. Trust Region Masking enforces this constraint at the sequence level, enabling the first non-vacuous monotonic improvement guarantees for long-horizon LLM-RL. We discuss length-neutral extensions in Appendix~\ref{app:length-neutral}.

\bibliography{ref}
\bibliographystyle{plainnat}

\newpage
\appendix
\onecolumn

\section{Details on Off-Policy Mismatch in LLM-RL}
\label{app:mismatch}

\subsection{Backend Discrepancies}

Modern LLM-RL pipelines use separate stacks for inference and training:

\begin{center}
\begin{tabular}{ll}
    \toprule
    \textbf{Inference (vLLM/SGLang)} & \textbf{Training (Megatron/FSDP)} \\
    \midrule
    PagedAttention & FlashAttention-2 \\
    FP8/INT8 KV-cache quantization & BF16/FP32 accumulation \\
    Aggressive operator fusion & Tensor parallelism \\
    \bottomrule
\end{tabular}
\end{center}

The root cause is floating-point non-associativity: $(a \oplus b) \oplus c \neq a \oplus (b \oplus c)$. Different parallel reduction orders in the softmax denominator yield slightly different results. While negligible for a single token, these errors compound autoregressively over $T$ steps.

\subsection{MoE Routing Discontinuities}

In MoE models~\citep{shazeer2017outrageously, liu2024deepseek}, the output is $y = \sum_{i \in \mathcal{K}} g_i(x) \cdot E_i(x)$ where $\mathcal{K} = \mathrm{Top\text{-}K}(h(x))$. The Top-K operator is discontinuous: if numerical jitter shifts $h_{\mathrm{inf}} = h_{\mathrm{train}} + \varepsilon$ such that $|h_{(K)} - h_{(K+1)}| < \|\varepsilon\|$, the selected experts change. This can cause $\piroll(v) = 0.9$ but $\pitheta(v) \approx 0.001$, producing importance ratio spikes $\rho \approx 900$.

\subsection{Distributed Staleness}

Actor-learner architectures~\citep{espeholt2018impala, nair2015massively} generate rollouts with $\theta_{\mathrm{old}}$ while the learner updates to $\theta_{\mathrm{new}}$. A lag of $k$ gradient steps ensures $\piroll \neq \pitheta$ even with identical implementations.

\section{Proofs of Foundational Lemmas}
\label{app:lemma-proofs}

\subsection{KL Chain Rule (Lemma~\ref{lem:kl-chain-app})}

\begin{lemma}[KL Chain Rule]
\label{lem:kl-chain-app}
For any time step $t$:
\begin{equation}
    \DKL(d_t^{\piroll} \| d_t^{\pitheta}) = \sum_{s=1}^{t-1} \E_{c_s \sim d_s^{\piroll}}[\DKL(c_s)].
\end{equation}
\end{lemma}

\begin{proof}
The joint trajectory distribution factorizes as:
\begin{equation}
    d_t^{\pi}(x, y_{<t}) = P(x) \prod_{s=1}^{t-1} \pi(y_s | c_s).
\end{equation}
Since $P(x)$ is shared, the KL divergence is:
\begin{align}
    \DKL(d_t^{\piroll} \| d_t^{\pitheta})
    &= \E_{(x, y_{<t}) \sim d_t^{\piroll}}\!\left[\log \frac{d_t^{\piroll}(x, y_{<t})}{d_t^{\pitheta}(x, y_{<t})}\right] \\
    &= \E_{d_t^{\piroll}}\!\left[\log \frac{P(x)\prod_{s=1}^{t-1}\piroll(y_s|c_s)}{P(x)\prod_{s=1}^{t-1}\pitheta(y_s|c_s)}\right] \\
    &= \E_{d_t^{\piroll}}\!\left[\sum_{s=1}^{t-1} \log \frac{\piroll(y_s|c_s)}{\pitheta(y_s|c_s)}\right] \\
    &= \sum_{s=1}^{t-1} \E_{d_t^{\piroll}}\!\left[\log \frac{\piroll(y_s|c_s)}{\pitheta(y_s|c_s)}\right]. \label{eq:kl-chain-step}
\end{align}
For each term in the sum, the expectation over $(x, y_{<t})$ under $d_t^{\piroll}$ can be decomposed. The log-ratio $\log(\piroll(y_s|c_s)/\pitheta(y_s|c_s))$ depends only on $(c_s, y_s) = (x, y_{\le s})$. Marginalizing out $y_{s+1}, \ldots, y_{t-1}$:
\begin{align}
    \E_{d_t^{\piroll}}\!\left[\log \frac{\piroll(y_s|c_s)}{\pitheta(y_s|c_s)}\right]
    &= \E_{(x, y_{\le s}) \sim d_{s+1}^{\piroll}}\!\left[\log \frac{\piroll(y_s|c_s)}{\pitheta(y_s|c_s)}\right] \\
    &= \E_{c_s \sim d_s^{\piroll}}\!\left[\E_{y_s \sim \piroll(\cdot|c_s)}\!\left[\log \frac{\piroll(y_s|c_s)}{\pitheta(y_s|c_s)}\right]\right] \\
    &= \E_{c_s \sim d_s^{\piroll}}[\DKL(c_s)].
\end{align}
Substituting back into Eq.~\eqref{eq:kl-chain-step} completes the proof.
\end{proof}

\begin{corollary}
\label{cor:kl-chain-bound}
$\DKL(d_t^{\piroll} \| d_t^{\pitheta}) \le (t-1)\,\delta$ and $\Dklseq = \sum_{t=1}^T \E[\DKL(c_t)] \le T\delta$.
\end{corollary}
\begin{proof}
Each summand satisfies $\E_{c_s}[\DKL(c_s)] \le \max_{c_s} \DKL(c_s) \le \delta$.
\end{proof}

\subsection{Martingale Property}

\begin{lemma}[Martingale Property]
\label{lem:martingale-app}
For any context $c_t$: $\E_{y_t \sim \piroll(\cdot|c_t)}[A_t^{\piroll}(c_t, y_t)] = 0$.
\end{lemma}

\begin{proof}
Define $Q_t(c_t, y_t) := \E_{\piroll}[R \mid c_t, y_t]$ and $V_t(c_t) := \E_{\piroll}[R \mid c_t] = \E_{y_t \sim \piroll}[Q_t(c_t, y_t)]$. Then $A_t = Q_t - V_t$, so:
\begin{equation}
    \E_{y_t \sim \piroll}[A_t(c_t, y_t)] = \E_{y_t \sim \piroll}[Q_t(c_t, y_t)] - V_t(c_t) = V_t(c_t) - V_t(c_t) = 0.
\end{equation}
\end{proof}

\subsection{Advantage Bound (Lemma~\ref{lem:advantage-bound})}

\begin{proof}[Full proof]
Using the Martingale Property:
\begin{align}
    g_t(c_t) &= \E_{y_t \sim \pitheta}[A_t(c_t, y_t)] - \underbrace{\E_{y_t \sim \piroll}[A_t(c_t, y_t)]}_{= 0} \\
    &= \sum_{y_t \in \mathcal{V}} \big(\pitheta(y_t|c_t) - \piroll(y_t|c_t)\big) \cdot A_t(c_t, y_t). \label{eq:gt-expansion}
\end{align}
\textbf{Bounding $|A_t|$:} Since $R \in [0,1]$, both $Q_t(c_t, y_t) = \E[R|c_t, y_t]$ and $V_t(c_t) = \E[R|c_t]$ lie in $[0,1]$. Therefore $|A_t| = |Q_t - V_t| \le 1$.

\textbf{H\"{o}lder's inequality:} Taking absolute values in Eq.~\eqref{eq:gt-expansion}:
\begin{align}
    |g_t(c_t)| &\le \sum_{y_t} |\pitheta(y_t|c_t) - \piroll(y_t|c_t)| \cdot |A_t(c_t, y_t)| \\
    &\le \sum_{y_t} |\pitheta(y_t|c_t) - \piroll(y_t|c_t)| \cdot 1 \\
    &= 2\,\Dtvtok(c_t). \label{eq:gt-tv-bound}
\end{align}

\textbf{Capping:} Since $\Dtvtok(c_t) \le 1$ for any pair of distributions, and by Pinsker's inequality $\Dtvtok(c_t) \le \sqrt{\DKL(c_t)/2}$:
\begin{equation}
    |g_t(c_t)| \le 2\min\!\big(1, \sqrt{\DKL(c_t)/2}\big).
\end{equation}
Maximizing over $c_t$: $\|g_t\|_\infty \le 2\min\!\big(1,\; \epsilon,\; \sqrt{\delta/2}\big)$.
\end{proof}

\subsection{Context Shift (Lemma~\ref{lem:context-shift})}

\begin{proof}[Full proof]
We establish each of the five bounds in Eq.~\eqref{eq:context-shift-all}.

\textbf{Bound 1: Trivial.} $\DTV(P, Q) \le 1$ for any distributions $P, Q$.

\textbf{Bound 2: Coupling.} We prove $\|d_t^{\pitheta} - d_t^{\piroll}\|_{\mathrm{TV}} \le (t-1)\,\epsilon$ by induction.

\emph{Base case ($t = 1$):} $d_1^{\pitheta} = d_1^{\piroll} = P(x)$, so $\DTV = 0 = (1-1)\epsilon$.

\emph{Inductive step:} Assume $\|d_t^{\pitheta} - d_t^{\piroll}\|_{\mathrm{TV}} \le (t-1)\epsilon$. The distribution at step $t+1$ is:
\begin{equation}
    d_{t+1}^{\pi}(x, y_{\le t}) = d_t^{\pi}(x, y_{<t}) \cdot \pi(y_t | c_t).
\end{equation}
Using the triangle inequality for product distributions (coupling lemma):
\begin{align}
    \|d_{t+1}^{\pitheta} - d_{t+1}^{\piroll}\|_{\mathrm{TV}}
    &= \frac{1}{2}\sum_{x, y_{\le t}} |d_t^{\pitheta}(c_t)\,\pitheta(y_t|c_t) - d_t^{\piroll}(c_t)\,\piroll(y_t|c_t)| \\
    &\le \frac{1}{2}\sum_{c_t, y_t} |d_t^{\pitheta}(c_t) - d_t^{\piroll}(c_t)| \cdot \pitheta(y_t|c_t) \nonumber\\
    &\quad + \frac{1}{2}\sum_{c_t, y_t} d_t^{\piroll}(c_t) \cdot |\pitheta(y_t|c_t) - \piroll(y_t|c_t)| \\
    &= \|d_t^{\pitheta} - d_t^{\piroll}\|_{\mathrm{TV}} \cdot \underbrace{\sum_{y_t}\pitheta(y_t|c_t)}_{=1} + \E_{c_t \sim d_t^{\piroll}}[\Dtvtok(c_t)] \\
    &\le (t-1)\epsilon + \epsilon = t\,\epsilon.
\end{align}
In the last step, we used the inductive hypothesis and $\Dtvtok(c_t) \le \epsilon$ for all $c_t$.

\textbf{Bound 3: Pinsker on marginal KL.} By the KL chain rule (\Cref{lem:kl-chain-app}):
\begin{equation}
    \DKL(d_t^{\piroll} \| d_t^{\pitheta}) = \sum_{s=1}^{t-1}\E_{c_s}[\DKL(c_s)] \le (t-1)\,\delta.
\end{equation}
Applying Pinsker's inequality to the marginal distributions:
\begin{equation}
    \|d_t^{\pitheta} - d_t^{\piroll}\|_{\mathrm{TV}} \le \sqrt{\frac{\DKL(d_t^{\piroll} \| d_t^{\pitheta})}{2}} \le \sqrt{\frac{(t-1)\,\delta}{2}}.
\end{equation}

\textbf{Bound 4: Data processing.} Since $d_t^{\pi}$ is a marginal of the full trajectory distribution $P^{\pi}(y|x)$ (obtained by marginalizing out $y_t, \ldots, y_T$), the data processing inequality gives:
\begin{equation}
    \|d_t^{\pitheta}(\cdot|x) - d_t^{\piroll}(\cdot|x)\|_{\mathrm{TV}} \le \|P^{\pitheta}(\cdot|x) - P^{\piroll}(\cdot|x)\|_{\mathrm{TV}} = \Dtvseq.
\end{equation}

\textbf{Bound 5: Pinsker on sequence KL.} Since $\DKL(d_t^{\piroll} \| d_t^{\pitheta}) \le \Dklseq$ (all summands in the chain rule are non-negative), Pinsker gives $\DTV \le \sqrt{\Dklseq/2}$.

Taking the minimum of all five bounds yields Eq.~\eqref{eq:context-shift-all}.
\end{proof}

\section{Proofs of Main Theorems}
\label{app:proofs}

All bounds start from the error decomposition (Eq.~\eqref{eq:error-pdi}):
\begin{equation}
    |\mathrm{Error}| \le \sum_{t=1}^T 2\|g_t\|_\infty \cdot \|d_t^{\pitheta} - d_t^{\piroll}\|_{\mathrm{TV}},
    \label{eq:master-decomp}
\end{equation}
which follows from $|\E_P[f] - \E_Q[f]| \le 2\|f\|_\infty \cdot \DTV(P, Q)$ applied to each summand.

\subsection{Pinsker-Marginal Bounds (\Cref{thm:pinsker-marginal})}

\textbf{KL variant.} Use the advantage bound $\|g_t\|_\infty \le 2\min(1, \sqrt{\delta/2})$ and the Pinsker context shift $\|d_t^{\pitheta} - d_t^{\piroll}\|_{\mathrm{TV}} \le \min(1, \sqrt{(t\!-\!1)\delta/2})$. Substituting into Eq.~\eqref{eq:master-decomp}:
\begin{align}
    |\mathrm{Error}| &\le \sum_{t=1}^T 2 \cdot 2\min\!\Big(1, \sqrt{\tfrac{\delta}{2}}\Big) \cdot \min\!\Big(1, \sqrt{\tfrac{(t-1)\delta}{2}}\Big) \\
    &= 4\min\!\Big(1, \sqrt{\tfrac{\delta}{2}}\Big) \sum_{t=1}^T \min\!\Big(1, \sqrt{\tfrac{(t-1)\delta}{2}}\Big) = B_{\mathrm{PM}}^{\mathrm{KL}}.
\end{align}

\textbf{Small-divergence simplification.} When $\delta \le 2/T$: for all $t \le T$, $(t\!-\!1)\delta/2 \le (T\!-\!1)\delta/2 < 1$, so the context-shift cap is inactive. Also $\delta/2 < 1/T < 1$, so the advantage cap is inactive. Then:
\begin{align}
    B_{\mathrm{PM}}^{\mathrm{KL}} &= 4\sqrt{\tfrac{\delta}{2}} \sum_{t=1}^T \sqrt{\tfrac{(t-1)\delta}{2}} = 4 \cdot \tfrac{\delta}{2} \sum_{k=0}^{T-1} \sqrt{k}.
\end{align}
The sum is bounded by the integral: $\sum_{k=0}^{T-1}\sqrt{k} \le \int_0^T \sqrt{x}\,dx = \frac{2}{3}T^{3/2}$. Therefore:
\begin{equation}
    B_{\mathrm{PM}}^{\mathrm{KL}} \le 4 \cdot \tfrac{\delta}{2} \cdot \tfrac{2}{3}T^{3/2} = \tfrac{4}{3}T^{3/2}\delta.
\end{equation}

\textbf{Verification at $T=4096$, $\delta=10^{-4}$:} $\frac{4}{3} \times (4096)^{3/2} \times 10^{-4} = \frac{4}{3} \times 262144 \times 10^{-4} = 34.95 \approx 35.0$.

\textbf{TV variant.} Replace the advantage bound with $\|g_t\|_\infty \le 2\min(1, \epsilon)$, keeping the Pinsker context shift unchanged:
\begin{equation}
    B_{\mathrm{PM}}^{\mathrm{TV}} = 4\min(1, \epsilon) \sum_{t=1}^T \min\!\Big(1, \sqrt{\tfrac{(t-1)\delta}{2}}\Big).
\end{equation}
In the small-divergence regime: $B_{\mathrm{PM}}^{\mathrm{TV}} = 4\epsilon \cdot \sqrt{\delta/2} \cdot \frac{2}{3}T^{3/2} = \frac{8}{3}T^{3/2}\epsilon\sqrt{\delta/2}$.

\textbf{Verification at $T=4096$, $\delta=10^{-4}$, $\epsilon=5\times10^{-3}$:}
$\frac{8}{3} \times 262144 \times 5\times10^{-3} \times \sqrt{5\times10^{-5}} = \frac{8}{3} \times 262144 \times 5\times10^{-3} \times 7.07\times10^{-3} = 24.7$.

Note: when $\epsilon = \sqrt{\delta/2}$ (Pinsker tight), $B_{\mathrm{PM}}^{\mathrm{TV}} = B_{\mathrm{PM}}^{\mathrm{KL}}$. The TV variant is strictly tighter only when $\epsilon < \sqrt{\delta/2}$.
\qed

\subsection{Mixed Bounds (\Cref{thm:mixed})}

\textbf{KL variant.} The marginal KL at any step $t$ is bounded by the full sequence KL:
\begin{equation}
    \DKL(d_t^{\piroll} \| d_t^{\pitheta}) = \sum_{s=1}^{t-1}\E[\DKL(c_s)] \le \sum_{s=1}^{T}\E[\DKL(c_s)] = \Dklseq.
\end{equation}
The inequality holds because all summands are non-negative. Applying Pinsker:
\begin{equation}
    \|d_t^{\pitheta} - d_t^{\piroll}\|_{\mathrm{TV}} \le \min\!\Big(1, \sqrt{\tfrac{\Dklseq}{2}}\Big).
    \label{eq:mixed-context-shift}
\end{equation}
This bound is uniform in $t$. Summing over $T$ steps with the KL advantage bound:
\begin{align}
    |\mathrm{Error}| &\le \sum_{t=1}^T 2 \cdot 2\min\!\Big(1, \sqrt{\tfrac{\delta}{2}}\Big) \cdot \min\!\Big(1, \sqrt{\tfrac{\Dklseq}{2}}\Big) \\
    &= 4T \cdot \min\!\Big(1, \sqrt{\tfrac{\delta}{2}}\Big) \cdot \min\!\Big(1, \sqrt{\tfrac{\Dklseq}{2}}\Big) = B_{\mathrm{Mix}}^{\mathrm{KL}}.
\end{align}
When both caps are inactive: $B_{\mathrm{Mix}}^{\mathrm{KL}} = 4T\sqrt{\delta/2}\sqrt{\Dklseq/2} = 2T\sqrt{\delta \cdot \Dklseq}$.

\textbf{Verification at $T=4096$, $\delta=10^{-4}$, $\Dklseq=0.01$:}
$2 \times 4096 \times \sqrt{10^{-4} \times 0.01} = 8192 \times 10^{-3} = 8.192 \approx 8.2$.

\textbf{TV variant.} Use the TV advantage bound $\|g_t\|_\infty \le 2\min(1, \epsilon)$ and the data-processing context shift $\|d_t^{\pitheta} - d_t^{\piroll}\|_{\mathrm{TV}} \le \min(1, \Dtvseq)$ (Bound 4 of \Cref{lem:context-shift}, proved in \Cref{app:lemma-proofs}):
\begin{equation}
    B_{\mathrm{Mix}}^{\mathrm{TV}} = 4T \cdot \min(1, \epsilon) \cdot \min(1, \Dtvseq).
\end{equation}

When caps are inactive: $B_{\mathrm{Mix}}^{\mathrm{TV}} = 4T\epsilon\,\Dtvseq$.

\textbf{Verification at $T=4096$, $\epsilon=5\times10^{-3}$, $\Dtvseq=0.05$:}
$4 \times 4096 \times 5\times10^{-3} \times 0.05 = 16384 \times 2.5\times10^{-4} = 4.096 \approx 4.1$.

When $\epsilon = \sqrt{\delta/2}$ and $\Dtvseq = \sqrt{\Dklseq/2}$: $B_{\mathrm{Mix}}^{\mathrm{TV}} = 4T\sqrt{\delta/2}\sqrt{\Dklseq/2} = 2T\sqrt{\delta\Dklseq} = B_{\mathrm{Mix}}^{\mathrm{KL}}$.
\qed

\subsection{Coupling Bound (\Cref{thm:coupling})}

Use the TV advantage bound $\|g_t\|_\infty \le 2\min(1, \epsilon)$ and the coupling context shift $\|d_t^{\pitheta} - d_t^{\piroll}\|_{\mathrm{TV}} \le \min(1, (t\!-\!1)\epsilon)$:
\begin{equation}
    |\mathrm{Error}| \le \sum_{t=1}^T 2 \cdot 2\min(1, \epsilon) \cdot \min(1, (t\!-\!1)\epsilon) = 4\min(1, \epsilon)\sum_{t=1}^T \min(1, (t\!-\!1)\epsilon) = B_{\mathrm{Coup}}.
\end{equation}

\textbf{Case 1: $T\epsilon \le 1$ (small divergence).} All caps are inactive:
\begin{equation}
    B_{\mathrm{Coup}} = 4\epsilon \cdot \epsilon\sum_{k=0}^{T-1}k = 4\epsilon^2 \cdot \tfrac{T(T-1)}{2} = 2T(T-1)\epsilon^2.
\end{equation}
This recovers the classical bound.

\textbf{Case 2: $T\epsilon > 1$ (large divergence).} Each summand satisfies $\min(1, (t\!-\!1)\epsilon) \le 1$, hence $\sum_{t=1}^T \min(1, (t\!-\!1)\epsilon) \le T$, giving $B_{\mathrm{Coup}} \le 4\min(1,\epsilon)\,T \le 4T\epsilon = O(T\epsilon)$.

\textbf{Verification at $T=4096$, $\epsilon=7.07\times10^{-3}$ (Pinsker-derived):} the cap activates at $t^\star = \lfloor 1/\epsilon \rfloor + 1 = 142$; evaluating $\sum_{t=1}^T \min(1, (t\!-\!1)\epsilon)$ exactly gives $\approx 4024$, so $B_{\mathrm{Coup}} = 4\epsilon \times 4024 \approx 113.8$, consistent with the bound $4T\epsilon = 115.9$.
\qed

\section{Proof of the Adaptive Bound (\Cref{thm:adaptive})}
\label{app:proof-adaptive}

The Adaptive bound uses an alternative error decomposition based on importance ratios rather than the Performance Difference Identity. We develop the proof in four steps.

\subsection{Step 1: Exact Error Identity}

\begin{lemma}[Exact Error Identity]
\label{lem:exact-identity}
\begin{equation}
    J(\pitheta) - J(\piroll) = L'_{\piroll}(\pitheta) - \Delta,
\end{equation}
where:
\begin{equation}
    \Delta := \E_{y \sim \piroll}\!\left[R(y)\sum_{t=1}^{T}\left(\rho_t - 1\right)\!\left(1 - \prod_{j=t+1}^{T}\rho_j\right)\right].
    \label{eq:delta-def}
\end{equation}
\end{lemma}

\begin{proof}
We use the telescoping identity for products. For any sequence $(\rho_1, \ldots, \rho_T)$:
\begin{equation}
    \prod_{t=1}^T \rho_t - 1 = \sum_{t=1}^T (\rho_t - 1)\prod_{j=t+1}^T \rho_j.
    \label{eq:telescoping}
\end{equation}
\emph{Proof of Eq.~\eqref{eq:telescoping}:} Write $\prod_{t=1}^T \rho_t = \rho_1 \cdot \prod_{j=2}^T \rho_j$. Then:
\begin{align}
    \prod_{t=1}^T \rho_t - 1 &= (\rho_1 - 1)\prod_{j=2}^T \rho_j + \Big(\prod_{j=2}^T \rho_j - 1\Big).
\end{align}
Applying the same decomposition recursively to $\prod_{j=2}^T \rho_j - 1$, and continuing, yields Eq.~\eqref{eq:telescoping}.

Now, the true performance difference is:
\begin{align}
    J(\pitheta) - J(\piroll)
    &= \E_{y \sim \piroll}\!\left[R(y)\left(\prod_{t=1}^T \rho_t - 1\right)\right] \\
    &= \E_{y \sim \piroll}\!\left[R(y)\sum_{t=1}^T (\rho_t - 1)\prod_{j>t}\rho_j\right]. \label{eq:Jdiff-expanded}
\end{align}
The surrogate is $L'_{\piroll}(\pitheta) = \E_{y \sim \piroll}[R(y)\sum_t (\rho_t - 1)]$. Subtracting:
\begin{align}
    J(\pitheta) - J(\piroll) - L'_{\piroll}(\pitheta)
    &= \E_{y \sim \piroll}\!\left[R(y)\sum_t (\rho_t - 1)\left(\prod_{j>t}\rho_j - 1\right)\right] \\
    &= -\E_{y \sim \piroll}\!\left[R(y)\sum_t (\rho_t - 1)\left(1 - \prod_{j>t}\rho_j\right)\right] = -\Delta.
\end{align}
Therefore $\mathrm{Error} = J(\pitheta) - J(\piroll) - L'(\pitheta) = -\Delta$, and $|\mathrm{Error}| = |\Delta|$.
\end{proof}

\subsection{Step 2: Tower Property Factorization}

Define for each $t$:
\begin{equation}
    A_t := |\rho_t - 1|, \qquad B_t := \left|1 - \prod_{j=t+1}^{T}\rho_j\right|.
\end{equation}
Since $|R(y)| \le 1$, the triangle inequality gives:
\begin{equation}
    |\Delta| \le \E_{y \sim \piroll}\!\left[\sum_{t=1}^{T} A_t \cdot B_t\right] = \sum_{t=1}^{T}\E_{y \sim \piroll}[A_t \cdot B_t].
    \label{eq:delta-triangle}
\end{equation}

The factor $A_t$ depends on $(c_t, y_t)$, while $B_t$ depends on $y_{>t} = (y_{t+1}, \ldots, y_T)$. They are \emph{not} independent given $c_t$ (since $c_{t+1} = (c_t, y_t)$ determines the future). Since $A_t$ is $y_{\le t}$-measurable (it depends only on $(c_t, y_t)$, and $c_{t+1} = (c_t, y_t)$), conditioning on $y_{\le t}$ and applying the tower property gives:
\begin{align}
    \E_{y \sim \piroll}[A_t \cdot B_t]
    &= \E_{y_{\le t} \sim \piroll}\!\left[A_t \cdot \E_{y_{>t} \sim \piroll(\cdot|c_{t+1})}[B_t]\right].
    \label{eq:tower-correct}
\end{align}

Now we compute the inner expectation. For fixed $c_{t+1}$:
\begin{align}
    \E_{y_{>t} \sim \piroll(\cdot|c_{t+1})}\![B_t]
    &= \E_{y_{>t} \sim \piroll}\!\left[\left|1 - \prod_{j=t+1}^T \frac{\pitheta(y_j|c_j)}{\piroll(y_j|c_j)}\right|\right] \\
    &= \E_{y_{>t} \sim \piroll}\!\left[\left|1 - \frac{P^{\pitheta}(y_{>t}|c_{t+1})}{P^{\piroll}(y_{>t}|c_{t+1})}\right|\right] \\
    &= \sum_{y_{>t}} P^{\piroll}(y_{>t}|c_{t+1}) \left|1 - \frac{P^{\pitheta}(y_{>t}|c_{t+1})}{P^{\piroll}(y_{>t}|c_{t+1})}\right| \\
    &= \sum_{y_{>t}} |P^{\piroll}(y_{>t}|c_{t+1}) - P^{\pitheta}(y_{>t}|c_{t+1})| \\
    &= 2\,\DTV(P^{\piroll}(\cdot|c_{t+1}) \| P^{\pitheta}(\cdot|c_{t+1})). \label{eq:Bt-is-TV}
\end{align}
This is the total variation distance between the future-trajectory distributions.

For the outer factor, we expand the expectation over $(c_t, y_t)$:
\begin{align}
    \E_{y_{\le t} \sim \piroll}[A_t] &= \E_{c_t \sim d_t^{\piroll}}\!\left[\E_{y_t \sim \piroll(\cdot|c_t)}\!\left[|\rho_t - 1|\right]\right].
\end{align}
We compute the inner expectation:
\begin{align}
    \E_{y_t \sim \piroll(\cdot|c_t)}\!\left[\left|\frac{\pitheta(y_t|c_t)}{\piroll(y_t|c_t)} - 1\right|\right]
    &= \sum_{y_t} \piroll(y_t|c_t) \left|\frac{\pitheta(y_t|c_t)}{\piroll(y_t|c_t)} - 1\right| \\
    &= \sum_{y_t} |\pitheta(y_t|c_t) - \piroll(y_t|c_t)| \\
    &= 2\,\Dtvtok(c_t). \label{eq:At-is-TV}
\end{align}
Therefore:
\begin{equation}
    \E_{y_{\le t}}[A_t] = 2\,\E_{c_t \sim d_t^{\piroll}}[\Dtvtok(c_t)] = 2\,\Dbar_t.
    \label{eq:At-expectation}
\end{equation}

\subsection{Step 3: Bounding the Future-Trajectory TV}

We bound $\DTV(P^{\piroll}(\cdot|c_{t+1}) \| P^{\pitheta}(\cdot|c_{t+1}))$ for any fixed $c_{t+1}$. The future trajectory has $T - t$ steps.

\textbf{Route A (Trivial):}
\begin{equation}
    \DTV(P^{\piroll}(\cdot|c_{t+1}) \| P^{\pitheta}(\cdot|c_{t+1})) \le 1.
    \label{eq:future-trivial}
\end{equation}

\textbf{Route B (Pinsker + KL chain rule):} By Pinsker's inequality applied to the future joint distribution:
\begin{equation}
    \DTV \le \sqrt{\frac{1}{2}\DKL(P^{\piroll}(\cdot|c_{t+1}) \| P^{\pitheta}(\cdot|c_{t+1}))}.
    \label{eq:future-pinsker}
\end{equation}
The conditional future KL decomposes via the chain rule:
\begin{align}
    \DKL(P^{\piroll}(\cdot|c_{t+1}) \| P^{\pitheta}(\cdot|c_{t+1}))
    &= \sum_{k=t+1}^{T}\E_{c_k \sim \piroll(\cdot|c_{t+1})}[\DKL(c_k)] \\
    &\le (T - t) \cdot \max_{k, c_k} \DKL(c_k) = (T-t)\,\delta. \label{eq:future-kl}
\end{align}
Combining: $\DTV \le \sqrt{(T-t)\delta/2}$.

\textbf{Route C (Coupling):} Applying the coupling argument to the $T-t$ remaining steps:
\begin{equation}
    \DTV(P^{\piroll}(\cdot|c_{t+1}) \| P^{\pitheta}(\cdot|c_{t+1})) \le (T-t)\,\epsilon.
    \label{eq:future-coupling}
\end{equation}
This follows from the same inductive argument as the context-shift coupling bound, applied to the conditional future distributions.

\textbf{Combined:} Taking the minimum of all three routes:
\begin{equation}
    \DTV(P^{\piroll}(\cdot|c_{t+1}) \| P^{\pitheta}(\cdot|c_{t+1})) \le \min\!\Big(1,\;(T\!-\!t)\,\epsilon,\;\sqrt{\tfrac{(T-t)\,\delta}{2}}\Big).
    \label{eq:future-combined}
\end{equation}

\textbf{Crucially}, this bound holds for \emph{every} realization of $c_{t+1}$, because the worst-case replacement in Eqs.~\eqref{eq:future-kl}--\eqref{eq:future-coupling} does not depend on the specific context. Therefore, the bound can be pulled outside the outer expectation over $c_{t+1}$.

\subsection{Step 4: Combining the Factors}

Substituting Eqs.~\eqref{eq:Bt-is-TV}, \eqref{eq:At-expectation}, and \eqref{eq:future-combined} into Eq.~\eqref{eq:tower-correct}:
\begin{align}
    |\Delta| &\le \sum_{t=1}^T \E_{y_{\le t}}[A_t] \cdot \sup_{c_{t+1}} 2\DTV(P^{\piroll}(\cdot|c_{t+1}) \| P^{\pitheta}(\cdot|c_{t+1})) \\
    &\le \sum_{t=1}^T 2\Dbar_t \cdot 2\min\!\Big(1,\;(T\!-\!t)\,\epsilon,\;\sqrt{\tfrac{(T-t)\delta}{2}}\Big) \\
    &= 4\sum_{t=1}^T \Dbar_t \cdot \min\!\Big(1,\;(T\!-\!t)\,\epsilon,\;\sqrt{\tfrac{(T-t)\delta}{2}}\Big) = B_{\mathrm{Adap}}^*.
\end{align}
This proves Eq.~\eqref{eq:B-Adap-star}. The KL and TV variants (Eqs.~\eqref{eq:B-Adap-KL}--\eqref{eq:B-Adap-TV}) follow by dropping one of the two non-trivial terms inside the $\min$.

\subsection{Step 5: Strictness over Prior Bounds}

\textbf{Recovery of $B_{\mathrm{PM}}^{\mathrm{KL}}$.} Set $\Dbar_t = \min(1, \sqrt{\delta/2})$ (worst-case per-position TV via Pinsker) and use only the Pinsker route for the future TV in $B_{\mathrm{Adap}}^{\mathrm{KL}}$:
\begin{equation}
    B_{\mathrm{Adap}}^{\mathrm{KL}} = 4\min\!\Big(1, \sqrt{\tfrac{\delta}{2}}\Big) \sum_{t=1}^T \min\!\Big(1, \sqrt{\tfrac{(T-t)\delta}{2}}\Big).
\end{equation}
By the index substitution $k = T - t$ (so $t = T - k$, and $k$ ranges from $0$ to $T-1$):
\begin{equation}
    \sum_{t=1}^T \min\!\Big(1, \sqrt{\tfrac{(T-t)\delta}{2}}\Big) = \sum_{k=0}^{T-1} \min\!\Big(1, \sqrt{\tfrac{k\delta}{2}}\Big) = \sum_{t=1}^T \min\!\Big(1, \sqrt{\tfrac{(t-1)\delta}{2}}\Big).
\end{equation}
This is exactly the sum in $B_{\mathrm{PM}}^{\mathrm{KL}}$ (Eq.~\eqref{eq:B-PM-KL}), confirming $B_{\mathrm{Adap}}^{\mathrm{KL}}|_{\Dbar_t = \sqrt{\delta/2}} = B_{\mathrm{PM}}^{\mathrm{KL}}$.

\textbf{Recovery of $B_{\mathrm{Coup}}$.} Set $\Dbar_t = \epsilon$ and use only the coupling route in $B_{\mathrm{Adap}}^{\mathrm{TV}}$:
\begin{equation}
    B_{\mathrm{Adap}}^{\mathrm{TV}} = 4\epsilon\sum_{t=1}^T\min(1, (T\!-\!t)\epsilon) = 4\epsilon\sum_{k=0}^{T-1}\min(1, k\epsilon) = B_{\mathrm{Coup}}.
\end{equation}

\textbf{Strict improvement.} Since $\Dbar_t \le \min(1, \epsilon, \sqrt{\delta/2})$ always, and strict inequality $\Dbar_t < \epsilon$ or $\Dbar_t < \sqrt{\delta/2}$ holds whenever the per-position divergence is non-uniform, the Adaptive bounds are strictly tighter than their PM/Coupling counterparts whenever divergence is non-uniform across positions.

Additionally, $B_{\mathrm{Adap}}^*$ with the per-position $\min$ over coupling and Pinsker routes is at least as tight as $\min(B_{\mathrm{Adap}}^{\mathrm{KL}}, B_{\mathrm{Adap}}^{\mathrm{TV}})$, since the per-position minimum is at most the global minimum of the two sums.
\qed

\subsection{Crossover Analysis}
\label{app:crossover}

The hybrid bound $B_{\mathrm{Adap}}^*$ selects between two routes for the future TV at each position. The crossover occurs where $(T\!-\!t)\epsilon = \sqrt{(T\!-\!t)\delta/2}$, i.e., $(T\!-\!t) = \delta/(2\epsilon^2)$.

\textbf{When Pinsker is tight} ($\epsilon = \sqrt{\delta/2}$): crossover at $T - t = 1$. The Pinsker route is tighter for all but the last position, and the hybrid reduces to $B_{\mathrm{Adap}}^{\mathrm{KL}}$.

\textbf{When Pinsker is loose} ($\epsilon \ll \sqrt{\delta/2}$): crossover at $T - t = \delta/(2\epsilon^2) \gg 1$. The coupling route wins for positions near the end (small $T\!-\!t$), while Pinsker wins for early positions (large $T\!-\!t$). The hybrid interpolates, providing significant improvement over either route alone.

\section{Sample-Based Estimators ($k_2$ and $k_3$)}
\label{app:k3}

When storing full logits is infeasible, we use sample-based estimators from $\rho_t = \pitheta(y_t|c_t)/\piroll(y_t|c_t)$.

\textbf{$k_3$ for averaging:} $f(\rho) = \rho - 1 - \log\rho$. This is the unique estimator satisfying $\E_{y_t \sim \piroll}[k_3(\rho_t)] = \DKL(\piroll(\cdot|c_t) \| \pitheta(\cdot|c_t))$ exactly, verified by:
\begin{align}
    \E_{\piroll}[\rho - 1 - \log\rho]
    &= \E_{\piroll}\!\Big[\frac{\pitheta}{\piroll} - 1\Big] - \E_{\piroll}\!\Big[\log\frac{\pitheta}{\piroll}\Big] = (1 - 1) + \E_{\piroll}\!\Big[\log\frac{\piroll}{\pitheta}\Big] = \DKL.
\end{align}
It is non-negative ($k_3 \ge 0$ by Jensen's inequality) and asymmetric (penalizes $\rho \gg 1$ more than $\rho \ll 1$).

\textbf{$k_2$ for max-filtering:} $f(\rho) = \frac{1}{2}(\log\rho)^2$. This is symmetric: $k_2(\rho) = k_2(1/\rho)$. It detects both support collapse ($\rho \to 0$) and impulse noise ($\rho \to \infty$) equally, which is essential for the max-based criterion. Note $k_2$ is biased as an estimator of $\DKL$ but serves as a robust outlier detector.

The rigorous guarantees of \Cref{thm:trm-guarantee} hold only with exact KL from full logits.

\section{Length-Neutral Trust Region Masking}
\label{app:length-neutral}

\subsection{Length Bias in TRM}

While TRM's threshold $\delta$ is length-invariant, the \emph{rejection probability} increases with $T$. Under a simplifying i.i.d.\ assumption with per-token violation probability $p$:
\begin{equation}
    \Pr[\text{accept}] = (1-p)^T \approx e^{-pT}.
\end{equation}
This decays exponentially in $T$, systematically rejecting long sequences. For reasoning tasks where correct solutions often require longer chains of thought, this bias is concerning.

\subsection{Length-Neutral TRM (LN-TRM)}

Motivated by the Adaptive bound, we define a trajectory-level error score:
\begin{equation}
    W(y) := \sum_{t=1}^{T}|\rho_t - 1| \cdot w_t, \qquad w_t := \min\!\Big(1,\;(T\!-\!t)\epsilon,\;\sqrt{\tfrac{(T-t)\delta}{2}}\Big),
\end{equation}
and normalize by the weight sum $Z(T) := \sum_{t=1}^T w_t$ (precomputable). The masking criterion is:
\begin{equation}
    M(y) = \mathbb{I}\!\Big[\widetilde{W}(y) := W(y)/Z(T) \le \delta_W\Big].
\end{equation}
Since $\widetilde{W}$ is a \emph{weighted average} of $|\rho_t - 1|$, it is approximately length-invariant.

\begin{proposition}[LN-TRM Guarantee]
\label{prop:ln-trm-guarantee}
If $\Dkltokmax \le \delta$ globally, then $|\mathrm{Error}| \le 2\,\E_{\piroll}[W(y)]$. For accepted trajectories, $W(y) \le \delta_W \cdot Z(T)$.
\end{proposition}
\begin{proof}
From the Adaptive bound proof (Step 4): $|\mathrm{Error}| \le 2\sum_t \E[|\rho_t - 1|] \cdot w_t = 2\,\E[W(y)]$ by linearity of expectation (since $w_t$ is deterministic). For accepted sequences, $\widetilde{W}(y) \le \delta_W$ gives $W(y) \le \delta_W \cdot Z(T)$.
\end{proof}

\subsection{Simplified Variant: Sequence-Error Ratio (SER)}

For minimal implementation overhead:
\begin{equation}
    W_{\mathrm{SER}}(y) := \frac{1}{T}\sum_{t=1}^{T}|\rho_t - 1|, \qquad M(y) = \mathbb{I}[W_{\mathrm{SER}}(y) \le \delta_{\mathrm{SER}}].
\end{equation}
This adds three lines to any GRPO/PPO implementation:
\begin{lstlisting}
W = mean(abs(rho - 1), dim=-1)  # per-sequence average
M = (W <= delta_ser).float()     # binary mask
loss = loss * M                  # mask rejected sequences
\end{lstlisting}
SER connects to the linear bound via $\E[|\rho_t - 1|] = 2\Dtvtok(c_t)$: controlling $W_{\mathrm{SER}} \le \delta$ bounds the per-trajectory error at rate $O(T\delta)$. Since $W_{\mathrm{SER}}$ is an average, its variance decreases as $O(1/\sqrt{T})$ by CLT, making SER mildly biased \emph{in favor of} longer sequences. We recommend $\delta_{\mathrm{SER}} \in [0.03, 0.10]$.

\begin{table}[htbp]
\centering
\caption{Length bias properties of masking methods.}
\label{tab:length-bias}
\begin{tabular}{llll}
    \toprule
    \textbf{Method} & \textbf{Criterion} & \textbf{Rejection scaling} & \textbf{Guarantee} \\
    \midrule
    TRM-Max & $\max_t \DKL(c_t) \le \delta$ & $1-(1-p)^T$ (exponential) & Exact \\
    TRM-Avg & $\frac{1}{T}\sum_t \DKL(c_t) \le \delta$ & $\approx$ constant & Weaker (avg $\ne$ max) \\
    LN-TRM & $\widetilde{W}(y) \le \delta_W$ & $\approx$ constant & Up to $Z(T)$ factor \\
    SER & $\frac{1}{T}\sum|\rho_t-1| \le \delta$ & $\approx$ constant & Via linear bound \\
    \bottomrule
\end{tabular}
\end{table}
\section{Precondition-Free Masked Guarantee}
\label{app:precond-free}

\Cref{rem:masking-role} shows that the unified bound $B^*$ cannot be transferred from the full surrogate $L$ to the masked surrogate $L_{\mathrm{masked}}$: a high acceptance rate does not bound $|L - L_{\mathrm{masked}}|$. A guarantee for the masked objective is nonetheless available directly, by decomposing the performance difference over the accepted and rejected regions and pricing rejection explicitly, without assuming the global precondition of \Cref{thm:trm-guarantee}.

\begin{proposition}[Masked guarantee via region decomposition]
\label{prop:masked-guarantee}
Fix $\pitheta$ and let $\mathcal{A}$ be any acceptance region, with mask $M = \mathbb{I}[(x,y) \in \mathcal{A}]$ and complement $\mathcal{R} := \mathcal{A}^c$; the TRM region $\{(x,y):\max_t \DKL(c_t)\le\delta\}$ is the canonical choice. For a set $\mathcal{S}$ write $J_{\mathcal{S}}(\pi) := \E_{x}\E_{y \sim \pi}[R(x,y)\,\mathbb{I}[(x,y)\in\mathcal{S}]]$, and define the masked surrogate
\[
    L'_{\mathrm{masked}}(\pitheta) := \E_{\piroll}\!\Big[M \cdot R \cdot \textstyle\sum_{t=1}^{T}(\rho_t - 1)\Big]
\]
in the $R$-form of \Cref{rem:surrogate-equiv}.\footnote{Unlike the unmasked case, a baseline $b$ contributes $-b\,\E_{\piroll}[M\sum_t(\rho_t-1)]$, which is \emph{not} $\pitheta$-independent once $M$ depends on $\pitheta$. The $R$-form and the advantage-form $L_{\mathrm{masked}}$ (and their gradients) coincide under the precondition $M\equiv1$; we state the result for the $R$-form.} Assume $\mathrm{supp}(\pitheta)\subseteq\mathrm{supp}(\piroll)$. Then
\begin{equation}
    J(\pitheta) - J(\piroll)
    = \underbrace{\big[J_{\mathcal{A}}(\pitheta) - J_{\mathcal{A}}(\piroll)\big]}_{(\mathrm{I})}
    + \underbrace{\big[J_{\mathcal{R}}(\pitheta) - J_{\mathcal{R}}(\piroll)\big]}_{(\mathrm{II})},
    \label{eq:region-decomp}
\end{equation}
where, with rejection rate $r := \E_{\piroll}[1 - M] = \Pr_{\piroll}[\mathcal{R}]$,
\begin{align}
    (\mathrm{I}) &= L'_{\mathrm{masked}}(\pitheta) - \Delta_{\mathcal{A}},
    \quad
    \Delta_{\mathcal{A}} := \E_{\piroll}\!\Big[M R \textstyle\sum_t (\rho_t - 1)\big(1 - \textstyle\prod_{j>t}\rho_j\big)\Big],
    \label{eq:accepted-term} \\
    |(\mathrm{II})| &\le r + \Dtvseq. \label{eq:rejected-penalty}
\end{align}
If in addition $|\Delta_{\mathcal{A}}| \le B_{\mathcal{A}}$ for some $B_{\mathcal{A}}$, then
\begin{equation}
    J(\pitheta) - J(\piroll) \;\ge\; L'_{\mathrm{masked}}(\pitheta) - B_{\mathcal{A}} - r - \Dtvseq,
    \label{eq:masked-guarantee}
\end{equation}
so $L'_{\mathrm{masked}}(\pitheta) > B_{\mathcal{A}} + r + \Dtvseq$ guarantees $J(\pitheta) > J(\piroll)$.
\end{proposition}

\begin{proof}
Eq.~\eqref{eq:region-decomp} is immediate from $\mathbb{I}_{\mathcal{A}} + \mathbb{I}_{\mathcal{R}} = 1$. For~$(\mathrm{II})$: since $R \in [0,1]$, $J_{\mathcal{R}}(\pi) \in [0, \Pr_{\pi}[\mathcal{R}]]$, so $-r \le (\mathrm{II}) \le \Pr_{\pitheta}[\mathcal{R}]$; as $|\Pr_{\pitheta}[\mathcal{R}] - \Pr_{\piroll}[\mathcal{R}]| \le \DTV(P^{\pitheta}, P^{\piroll}) = \Dtvseq$, we have $\Pr_{\pitheta}[\mathcal{R}] \le r + \Dtvseq$, giving Eq.~\eqref{eq:rejected-penalty}. For~$(\mathrm{I})$: by importance sampling on $\mathcal{A}$ (shared support), $J_{\mathcal{A}}(\pitheta) - J_{\mathcal{A}}(\piroll) = \E_{\piroll}[M R(\prod_t \rho_t - 1)]$; the telescoping identity $\prod_t\rho_t - 1 = \sum_t(\rho_t-1) - \sum_t(\rho_t-1)(1-\prod_{j>t}\rho_j)$ yields Eq.~\eqref{eq:accepted-term}. Eq.~\eqref{eq:masked-guarantee} combines the three displays via $J(\pitheta) - J(\piroll) \ge (\mathrm{I}) - |(\mathrm{II})|$ and $(\mathrm{I}) \ge L'_{\mathrm{masked}} - |\Delta_{\mathcal{A}}|$.
\end{proof}

\Cref{prop:masked-guarantee} bounds the \emph{value} of the $R$-form surrogate $L'_{\mathrm{masked}}$; this coincides with the advantage-form objective that Algorithm~\ref{alg:trm} ascends only when $M \equiv 1$ (\Cref{rem:surrogate-equiv}). In the precondition-free regime ($M$ varying) the result therefore characterizes what masking can certify in principle, rather than the specific stop-gradient update.

\begin{remark}[On the accepted-region bound $B_{\mathcal{A}}$]
\label{rem:BA-caveat}
The penalty $r + \Dtvseq$ in Eq.~\eqref{eq:rejected-penalty} is unconditional ($\Dtvseq \le \sqrt{\Dklseq/2}$, and $r$ is exactly the empirical masking rate); only $\Delta_{\mathcal{A}}$ requires control. This is the conditional step because $M$ is a \emph{trajectory-level} event: the per-position argument of \Cref{thm:adaptive} bounds the future factor $|1 - \prod_{j>t}\rho_j|$ only when the $\piroll$-reachable futures of $c_{t+1}$ stay within the trust region, which acceptance of a \emph{prefix} need not certify. Two choices of $\mathcal{A}$ discharge it:
\begin{enumerate}
    \item \textbf{TRM region} ($\mathcal{A} = \{\max_t \DKL(c_t) \le \delta\}$). Under the global precondition $\Dkltokmax \le \delta$ and $\mathrm{supp}(\pitheta)\subseteq\mathrm{supp}(\piroll)$, every $\pitheta$-trajectory lies in $\mathcal{A}$, so $r = \Pr_{\pitheta}[\mathcal{R}] = 0$ and $(\mathrm{II}) = 0$, while $|\Delta_{\mathcal{A}}| = |\Delta| \le B^*$. The bound then recovers \Cref{thm:trm-guarantee}(3) \emph{exactly} (the case $B_{\mathcal{A}} = B^*$, with no residual slack).
    \item \textbf{Sample-ratio region} ($\mathcal{A} = \{\max_t |\log \rho_t| \le \tau\}$; cf.\ Appendix~\ref{app:k3}). Accepted trajectories have pointwise-bounded ratios, so $|1 - \prod_{j>t}\rho_j| \le e^{(T-t)\tau} - 1$ gives a bound $B_{\mathcal{A}}$ \emph{without} any global precondition. Here $r > 0$ is genuine and observable---the regime in which Eq.~\eqref{eq:masked-guarantee} adds content over \Cref{thm:trm-guarantee}, at the cost of an exponential dependence on $\tau$ that motivates small per-token thresholds. The exponential is largely intrinsic: because $M$ couples the entire trajectory, the tower-property factorization behind \Cref{thm:adaptive} no longer applies, and a sublinear future bound would require controlling higher moments of $\rho$.
\end{enumerate}
The rejection rate is thus the price of a precondition-free statement: $r$ enters linearly and observably, formalizing the recommendation to monitor the masking rate.
\end{remark}

\end{document}